\documentclass[sigconf, nonacm]{acmart}

\newcommand\vldbdoi{XX.XX/XXX.XX}
\newcommand\vldbpages{XXX-XXX}
\newcommand\vldbvolume{14}
\newcommand\vldbissue{1}
\newcommand\vldbyear{2020}
\newcommand\vldbauthors{\authors}
\newcommand\vldbtitle{\shorttitle} 
\newcommand\vldbavailabilityurl{http://vldb.org/pvldb/format_vol14.html}

\usepackage{color}
\usepackage{xcolor}
\usepackage{mdframed}
\usepackage{booktabs}
\usepackage[ruled,vlined]{algorithm2e}
\usepackage{algorithmic}
\usepackage{graphicx}
\usepackage{lineno}

\usepackage{multirow}
\usepackage{subcaption}
\newcommand{\neuraldatabases}{Neural Databases}

\newcommand{\ndb}{{\sc NeuralDB}}
\newcommand{\ndbtitle}{{NeuralDB}}
\newcommand{\systemname}{{NeuralDB}}
 
\newcommand{\tfidf}{{TF$\cdot$IDF}}

\newcommand{\ucam}{\affiliation{%
  \institution{University of Cambridge}
}}

\newcommand{\fbuk}{\affiliation{%
  \institution{Facebook AI}
}}

\newcommand{\fbus}{\affiliation{%
  \institution{Facebook AI}
}}

\author{James Thorne}%\footnote{Work completed during an internship at FB}}
\ucam
\fbuk
\email{jt719@cam.ac.uk}

\author{Majid Yazdani}
\fbuk
\email{myazdani@fb.com}

\author{Marzieh Saeidi}
\fbuk
\email{marzieh@fb.com}

\author{Fabrizio Silvestri}
\fbuk
\email{fsilvestri@fb.com}

\author{Sebastian Riedel}
\fbuk
\email{sriedel@fb.com}

\author{Alon Halevy}
\fbus
\email{ayh@fb.com}

\newcounter{jtCounter}
\newif\ifjtvar
\jtvarfalse
\ifjtvar
\newcommand{\jt}[1]{{\small \color{purple} \refstepcounter{jtCounter}\textsf{[JT]$_{\arabic{jtCounter}}$:{#1}}}}
\else
\newcommand{\jt}[1]{}
\fi

\newcounter{fsCounter}
\newif\iffsvar
\fsvarfalse
\iffsvar
\newcommand{\fs}[1]{{\small \color{orange} \refstepcounter{fsCounter}\textsf{[FS]$_{\arabic{fsCounter}}$:{#1}}}}
\else
\newcommand{\fs}[1]{}
\fi

\newcounter{ahCounter}
\newif\ifahvar
\ahvarfalse

\ifahvar
\newcommand{\ah}[1]{{\small \color{red} \refstepcounter{ahCounter}\textsf{[AH]$_{\arabic{ahCounter}}$:{#1}}}}
\else
\newcommand{\ah}[1]{}
\fi

\newcounter{srCounter}
\newif\ifsrvar
\srvarfalse
\ifsrvar
\newcommand{\sr}[1]{{\small \color{blue} \refstepcounter{srCounter}\textsf{[SR]$_{\arabic{srCounter}}$:{#1}}}}
\else
\newcommand{\sr}[1]{}
\fi

\newcounter{msCounter}
\newif\ifmsvar
\msvarfalse
\ifmsvar
\newcommand{\ms}[1]{{\small \color{cyan} \refstepcounter{msCounter}\textsf{[MS]$_{\arabic{msCounter}}$:{#1}}}}
\else
\newcommand{\ms}[1]{}
\fi

\newcounter{myCounter}
\newif\ifmyvar
\myvarfalse
\ifmyvar
\newcommand{\my}[1]{{\small \color{magenta} \refstepcounter{myCounter}\textsf{[MY]$_{\arabic{myCounter}}$:{#1}}}}
\else
\newcommand{\my}[1]{}
\fi

\begin{document}

% TODO Before final submisison remove page numvers
%\pagenumbering{arabic}
%\pagestyle{plain}
%\setcounter{page}{1}

\title{Neural Databases}
\begin{abstract}
\jt{TODO Before final submission remove page numbers}
In recent years, neural networks have shown impressive performance gains on long-standing AI problems, and in particular, answering queries from natural language text. These advances raise the question of whether they can be extended to a point where we can relax the fundamental assumption of database management, namely, that our data is represented as fields of a pre-defined schema. 

This paper presents a first step in answering that question. 
We describe \ndb, a database system with no pre-defined schema, in which updates and queries are given in natural language. We develop query processing techniques that build on the  primitives offered by the state of the art Natural Language Processing methods. 

We begin by demonstrating that at the core, recent NLP transformers, powered by pre-trained language models, can answer select-project-join queries if they are given the exact set of relevant facts. However, they cannot scale to non-trivial databases and cannot perform aggregation queries. Based on these findings, we describe a \ndb\ architecture that runs multiple Neural SPJ operators in parallel, each with a set of database sentences that can produce one of the answers to the query. The result of these operators is fed to an aggregation operator if needed. We describe an algorithm that learns how to create the appropriate sets of facts to be fed into each of the Neural SPJ operators. Importantly, this algorithm can be trained by the Neural SPJ operator itself. We experimentally validate the accuracy of \systemname\ and its components, showing that we can answer queries over thousands of sentences with very high accuracy. 
\end{abstract}

\maketitle

% See old.tex for the original doc

\begingroup\small\noindent\raggedright\textbf{PVLDB Reference Format:}\\
\vldbauthors. \vldbtitle. PVLDB, \vldbvolume(\vldbissue): \vldbpages, \vldbyear.\\
\href{https://doi.org/\vldbdoi}{doi:\vldbdoi}
\endgroup
\begingroup
\renewcommand\thefootnote{}\footnote{\noindent
This work is licensed under the Creative Commons BY-NC-ND 4.0 International License. Visit \url{https://creativecommons.org/licenses/by-nc-nd/4.0/} to view a copy of this license. For any use beyond those covered by this license, obtain permission by emailing \href{mailto:info@vldb.org}{info@vldb.org}. Copyright is held by the owner/author(s). Publication rights licensed to the VLDB Endowment. \\
\raggedright Proceedings of the VLDB Endowment, Vol. \vldbvolume, No. \vldbissue\ %
ISSN 2150-8097. \\
\href{https://doi.org/\vldbdoi}{doi:\vldbdoi} \\
}\addtocounter{footnote}{-1}\endgroup

\ifdefempty{\vldbavailabilityurl}{}{
\vspace{.3cm}
\begingroup\small\noindent\raggedright\textbf{PVLDB Availability Tag:}\\
The source code of this research paper has been made publicly available at \url{\vldbavailabilityurl}.
\endgroup
}

%\linenumbers

\section{Introduction}

In recent years, neural networks have shown impressive performance gains on long-standing AI problems, such as natural language understanding, speech recognition, and computer vision. 
Based on these successes, researchers have considered the application of neural nets to data management problems, including learning indices~\cite{DBLP:conf/sigmod/KraskaBCDP18}, query optimization and entity matching~\cite{DBLP:conf/sigmod/MudgalLRDPKDAR18,DBLP:journals/corr/abs-2004-00584}. 
In applying neural nets to data management, research has so far assumed that the data was modeled by a database schema.

The success of neural networks in processing unstructured data such as natural language and images   raises the question of whether their use can be extended to a point where we can relax the fundamental assumption of database management, which is that the data we process is represented as fields of a pre-defined schema.  What if, instead, data and queries can be represented as short natural language sentences, and queries can be answered from these sentences? 
This paper presents a first step in answering that question. 
We describe \systemname, a database system in which updates and queries are given in natural language. The query processor of a \ndb\ builds on the primitives that are offered by the state of the art Natural Language Processing~(NLP) techniques.  Figure~\ref{fig:actual_data} shows example facts and queries that \ndb\ can answer. %\ms{In Figure 1, queries 4&5 are not really joins, they just need language understanding/paraphrasing}

Realizing the vision of \systemname\ will offer several benefits that database systems have struggled to support for decades. 
The first, and most important benefit is that a \ndb, by definition, has no pre-defined schema.
Therefore, the scope of the database does not need to be defined in advance and any data that becomes relevant as the application is used can be stored and queried.
The second benefit is that updates and queries can be posed in a variety of natural language forms, as is convenient to any user. 
In contrast, a traditional database query needs to be based on the database schema.  A third benefit comes from the fact that the \ndb\  is based on a pre-trained language model that already contains a lot of knowledge.  
For example, the fact that London is in the UK is already encoded in the language model. Hence, a query asking who lives in the UK can retrieve people who are known to live in London without having to explicitly specify an additional join. Furthermore, using the same paradigm, we can endow the \ndb\  with more domain knowledge by extending the pre-training corpus to that domain.

By nature, a \ndb\ is not meant to provide the same correctness guarantees of a traditional database system, i.e., that the answers returned for a query satisfy the precise binary semantics of the query language. 
Hence, \ndb s should not be considered as an alternative to traditional databases in applications where such guarantees are required.

Given its benefits, \neuraldatabases\ are well suited for emerging applications where the schema of the data cannot be determined in advance and data can be stated in a wide range of linguistic patterns. 
A family of such applications arise in the area of storing knowledge for personal assistants that currently available for home use and in the future will accompany Augmented Reality glasses. In these applications, users store data about their habits and experiences, their friends and their preferences, and designing a schema for such an application is impractical. 
Another class of applications is the modeling and querying of political claims~\cite{Thorne2018a} (with the goal of verifying their correctness). 
Here too, claims can be about a huge variety of topics and expressed in many ways. 

Our first contribution is to show that state of the art transformer models~\cite{Vaswani2017} can be adapted to answer simple natural language queries. Specifically, the models can process facts that are relevant to a query independent of their specific linguistic form, and combine multiple facts to yield correct answers, effectively performing a join. However, we identify two major limitations of these models: (1) they do not perform well on aggregation queries (e.g., counting, max/min), and (2) since the input size to the transformer is bounded and the complexity of the transformer is quadratic in the size of its input, they only work on a relatively small collection of facts.

Our second contribution is to  propose an architecture for neural databases that uses the power of transformers at its core, but puts in place several other components in order to address the scalability and aggregation issues. Our architecture runs multiple instances of a Neural SPJ operator in parallel. The results of the operator are either the answer to the query or the input to an aggregation operator, which is done in a traditional fashion. Underlying this architecture is a novel algorithm for generating the small sets of database sentences that are fed to each Neural SPJ operator.

Finally, we describe an experimental study that validates the different components of \systemname s, namely the ability of the Neural SPJ to answer queries or create results for a subsequent aggregation operator even with minimal supervision, and our ability to produce support sets that are fed into each of the Neural SPJ operators. Putting all the components together, our 
 final result shows that we can accurately answer queries over thousands of sentences with very high accuracy. To run the experiments we had to create an experimental dataset with training data for \ndb s, which we make available for future research.

% and capable of generating intermediate results and (3) accurately predicting the aggregation operation to execute over these intermediate results. 

\section{Problem Definition}

\begin{figure}[t]
    \centering
\noindent
\begin{mdframed}
\textbf{Facts}: (4 of 50 shown)
\begin{tabbing}
Nicholas lives in Washington D.C. with Sheryl. \\
Sheryl is Nicholas's spouse. \\
Teuvo was born in 1912 in Ruskala. \\
In 1978, Sheryl's mother gave birth to her in Huntsville.
\end{tabbing}
\textbf{Queries:}
\begin{tabbing}
Does Nicholas's spouse live in Washington D.C.?\\
\texttt{(Boolean Join)} $\longrightarrow$ {\tt TRUE} \\ \\
Who is Sheryl's husband?  \\
\texttt{(Lookup)} $\longrightarrow$ Nicholas \\ \\
Who is the oldest person in the database? \\
\texttt{(Max)} $\longrightarrow$ Teuvo \\ \\
Who is Sheryl's mother?  \\
\texttt{(Lookup)} $\longrightarrow$ {\tt NULL}
\end{tabbing}

\end{mdframed}
    \caption{In \systemname, facts and queries are posed with short natural language sentences. The queries above are answered by our first prototype described in Section~\ref{section:first-experiment}. }
    \label{fig:actual_data}
\end{figure}

The main goal of \ndb\ is to support data management applications where users do not need to pre-define a schema. Instead, they can express the facts in the database in any linguistic form they want, and queries can be posed in natural language.  To that end, data and queries in a \ndb\ are represented as short sentences in natural language and the neural machinery of the \ndb\ is applied to these sentences. 

\medskip
\noindent
{\bf Data:}  
Users input data into an \ndb\ using simple natural language sentences. 
Intuitively, a sentence corresponds to a single fact, such as  {\sf Sue is Mary's mom}, or {\sf Gustavo likes espresso}. But in many situations, especially when updates to the database are spoken by users, it is more convenient for sentences to express multiple facts, such as {\sf Kiara is married to Kyrone and they have 3 kids}. We refer to the latter sentences as composite and the former as atomic. To focus on the novel issues of \ndb s, we mostly consider atomic sentences in this paper, but we do demonstrate in  Section~\ref{section:first-experiment}  both atomic and composite sentences.

Formally, the data in an \ndb\ is a set of sentences, each with a time stamp, $D = \{(u_1, t_1), \ldots, (u_k, t_k)\}$. A \ndb\ allows updates, deletions and queries. The user does not need to distinguish between an update and a query because NLP technology can reliably make that distinction.  We assume that deletions are explicitly marked and also refer to individual sentences. 

\medskip
\noindent
{\bf Queries:} Formally, a query $Q$ over a database, $D$, produces a set of answers: $Q(D) = \{a_1,\ldots,a_l\}$. While queries are formulated in natural language, we only consider queries that, if translated to SQL, would involve a select-project-join (SPJ) component followed by an aggregation. However, for our analysis we need to make some further distinctions between several classes of queries (outlined in Figure~\ref{fig:actual_data}). 
 
A {\em lookup query} is a query where each answer comes from a single fact  in the database (e.g., {\sf Who is Susan's husband?}), whether there is a single answer or several.  If the query returns True/False, we refer to it as a Boolean query. Note that in our context, lookup queries are non-trivial because facts in the database are expressed in a variety of linguistic forms.
% For example, facts about taking a vacation, flying to a destination, visiting friends in a city or driving across a border could all correspond to a visit to Scotland.  

A {\em join query} is one that requires combining two (or more) facts in the database in order to produce each answer. For example, the query {\sf Who works in a company in France?} could combine facts pertaining to a person's country of residence, and others pertaining to people's jobs. In some cases, the query may require a join even if one is not explicitly specified in the query. For example,  the query {\sf Who is John's uncle?} could involve combining a fact about John's parents and about their brothers.    

We also consider queries that require performing an aggregation (e.g., {\sf how many kids does Pat have?}). In this paper, we focus on the {\tt count}, {\tt min} and {\tt max} aggregation operators. 

We note that while there is an intuitive mapping between natural language queries and SQL constructs, the mapping is not always precise in the context of \ndb s. For example, even though a query is a mere lookup, it may involve an implicit join. Consider the query {\sf does Susan live in the UK?}, and a database that contains  {\em Susan lives in London}. In this case, the query answering engine of the \ndb\ will benefit from a rich underlying language model in order to deduce that London is in the UK, and therefore answer that Susan lives in the UK. Finally, the selections in our queries are equality predicates on strings. Comparison predicates will be handled in future work.

%    .... \ah{we should distinguish between the case in which the join is actually explicit in the query, such as John's wife's boss vs.\ the case in which a simple query, John's uncle, requires a join.}

% \medskip
% \noindent
% {\bf Computation:} \jt{should we describe computations we'd like to do with the query results too? We might have to perform an inference for Boolean Question Answering, or aggregate results with min/max/count etc}
% The query may require a computation or aggregation over facts stored in the database rather than merely extracting information from the facts. We consider support for aggregation such as {\tt min}, {\tt max} and {\tt count} as well as questions posed in natural language with Boolean answers such as {\sf Is John a barber?} or {\sf Does John have any brothers?} which require the model to perform reasoning over the available information.

\medskip
\noindent
{\bf Unique names assumption:} In order to focus on the new issues that are raised by \ndb s, in this paper we assume that a given name refers to exactly one entity in the domain and each entity has a single name. We also assume that pronouns (e.g., she, they) are not used or have been resolved in advance. 
The application of the rich body of work on entity resolution to \ndb s will be reserved for future work. %That being said, we do show that \ndb s are capable of handling a range of expressions with multiple ways of expressing the same fact: in Section~\ref{section:experiments:alias} we will demonstrate that aliases for entities and properties can be resolved using parametric information captured within pre-trained language models which warrants further investigation. 

\subsection{NLP background}  
\label{section:transformers}

The field of natural language processing has made impressive progress in recent years by building on transformer architectures and pre-trained language models. Such models have led to major improvements on tasks such as question answering from text, text summarization, and machine translation. 
In \ndb, our goal is to represent the facts in a database as a set of simple natural language sentences. Hence, as a first step to realizing \ndb s, we consider whether techniques for question answering on text can be adapted to our context. 

In this section we describe the relevant aspects of pre-trained language models and transformers that operate over them, and the challenges in adapting these techniques to our context.  

\subsubsection*{Pre-trained language models and transformers}
\label{section:explain-transformers}

Pre-trained language models such as {BERT}~\cite{Devlin2019}, {GPT-3}~\cite{Brown2020}, {RoBERTa}~\cite{Liu2019RobustlyOptimized}, {T5}~\cite{Raffel2020} are neural models trained on large corpora of text. The models are trained by randomly removing certain tokens from the corpus and training the neural net to predict the missing token, given its context (i.e., the words preceding and following the missing token). At an intuitive level, these models obtain two kinds of knowledge (1) the ability to make predictions independent of the exact linguistic form in which knowledge is stated~\cite{Peters2018, Tenney2019}, and (2) some world knowledge that is mentioned frequently in text (e.g., London is the capital of the UK)~\cite{Petroni2019, Brown2020}.\footnote{Note that the second type of knowledge can also lead to implicit biases. We address this point in Section~\ref{section:conclusions}} %\ms{should we mention common sense reasoning?} \ah{commonsense reasoning is a very loaded term. I would stay away unless we have a specific example we want to show.}
Pre-trained language models are usually fine-tuned to a given task. For example, for question answering from text, the system would be further trained with a pair of (question, answer) pairs in order to produce the final model. Importantly, since the pre-trained language models capture world knowledge~\cite{Petroni2019} and act as an implicit regularizer~\cite{Radford2018, Hao2019},  fewer examples are needed for the fine-tuning compared to training a model from scratch.

%OLD Pre-trained language models are usually fine-tuned to a given task. For example, for question answering from text, the system would be further trained with a pair of (question, answer) pairs in order to produce the final model. Importantly, since the pre-trained language model already has quite a bit of language knowledge, the number of examples needed in the fine-tuning step is much smaller than it would be if we trained a model from scratch. 

The Transformer model~\cite{Vaswani2017} is the most common neural architecture to operate on pre-trained language models based on the high accuracy it produces on downstream tasks including question answering on text. 
In our prototype experiments, detailed in Section~\ref{section:first-experiment}, we demonstrate that these reasoning abilities enable the transformer architecture to generate correct answers to a number of queries that we might pose to a \ndb.

\subsubsection*{Training models}
We will train the parameters of our neural network of the \ndb\ with examples that includes  (query, answer) pairs. 
In Section~\ref{section:datasets} we describe how we obtain training data by leveraging facts from Wikidata.
The training data sets we use contain in the order of 10,000-100,000 examples. In a sense, one can view the need for training data as the cost we pay to support a database with no pre-defined schema. Indeed, we will show that without training data, the performance of \systemname\ degrades quite a bit. However, training is important not only for understanding new relations, but also to be able to handle the different linguistic variations in which facts are expressed. In contrast, a schema will always be brittle and allow only one way of expressing a given fact. Furthermore, training data is more easily transferred between different applications. 

Along these lines, there is a question of
 whether the training data needs to cover all the relationships that are seen in the queries. For example, can a \ndb\ answer a query {\sf what are Ruth's hobbies?} if it has not been trained on examples that include mentions of hobbies. In Section~\ref{section:experiments} we show that while accuracy of answers drops for such {\em out of domain} queries, the number of additional training examples that need to be added to achieve reasonable performance is relatively small.

% Where neural networks are trained for the \ndb\, the supervised training data is generated through a controlled process using 
% For language model pre-training \cite{glove,word2vec} \cite{Peters2018, Devlin2019, Raffel2020}, supervision is readily available through masking tokens from sentences and comparing the predicted token against the masked token when computing the loss function. Because generating this supervised data can be generated from scraped web texts, it is common for language model pre-training to use billions of instances.

In more detail, training neural networks is an optimization problem: minimizing the expected loss over the instances in the training data through changes to the model parameters, weighted by their contribution to the error term over all instances in the training dataset. Formally,
the loss is the cross-entropy between the model prediction and the reference: $L = -\sum_{c\in C} p(y_c)\log \hat{p}({y}_c)$ which is non-negative and real-valued. 
%the non-negative real-valued loss function is denoted by $L(\hat{y},y)$, where $\hat{y}$ is the output of the model and $y$ is the value for the corresponding training example.  
For both sentence classification (assigning a single discrete label $c\in C$ for a sequence of tokens) and language generation tasks (decoding a sequence of tokens $(c_1,\ldots,c_n)$ from the vocabulary $C$ as output from the model) the model output is a probability distribution over labels $\hat{p}(y_c) \in C$. %For language generation, the cross-entropy loss is summed over all tokens.

% The model parameters that are fit on the training dataset describe a mapping from input to output. For the model to generalize, this mapping must also hold for data that the model has not been exposed to during training. %We measure model whether the model is overfitting the training data 
% We use held-out validation data and test data to estimate of generalization performance of the model to unseen data. A commonly held viewpoint is that language model pre-training aids model generalization by acting as an implicit regularizer \cite{Radford2018} which yields flatter optima during training \cite{Hao2019}. Furthermore, language model pre-training captures common sense and world knowledge that can be exploited by the model \cite{Petroni2019}.% making the model more robust to overfitting.

\subsubsection*{Evaluating accuracy of answers}
We measure the correctness of the answers generated by a \ndb\ by comparing them against reference data that contain the correct answers. The neural networks are trained with subset of the available data, leaving a portion of it held-out for evaluation, referred to as the test set.

% referred to as the test set. 
% To evaluate the accuracy of a \ndb, we 
% Evaluating the answers of NLP models and consequently of a \ndb\ is different than a 
% The training of neural networks, such as the transformer models in this paper, is a stochastic process: differences in training conditions will lead to changes in model parameters and variations in the answers generated by the \ndb. In our experiments, 

For most queries, we measure correctness using Exact Match (EM), which is~1 if  a binary  the answer string generated by the \ndb\ is exactly equal to the reference answer and~0 otherwise.  This metric is used to score outputs where either a Boolean, null answer, string or numeric answer is expected.

When a set of results is returned, we also consider the $F_1$ score that weighs the precision and recall of the answer generated by the \ndb\  as compared to the reference data. Precision and recall penalize false positives (denoted fp) and false negatives (denoted fn) compared to true positives (denoted tp) respectively: $p = \frac{tp}{tp+fp}$, $r = \frac{tp}{tp+fn}$. The $F_1$ score, ranging from 1.0 for perfect match to 0.0 for no match is the harmonic mean of precision and recall: $F_1 = \frac{2pr}{p+r}$ which has an affinity for the lower of two scores.
When comparing models and reporting results, we average the instance-level scores (either EM and $F_1$) over all examples in the test set.

% \medskip
% \noindent
% Additional diagnostic metrics for evaluating internal components of the \ndb\ - will be discussed as and when they are used.   
%We will not evaluate external aggregation modules (such as count, min/max etc) as these are deterministic functions that do not depend on training conditions. However, as part of the aggregation module, the appropriate aggregation operator must be selected. This is modelled as a classification task over the query, choosing an operator depending on the label predicted by this classifier. We report classification accuracy for this.

\subsubsection*{The transformer architecture}

Transformers~\cite{Vaswani2017} take as input an sequence of symbols $\mathbf{x}=(x_1,\ldots,x_n)$.
They are typically trained in one of two configurations: encoder only or encoder-decoder. In the former, each token is encoded to a vector representation that is used to predict a label. In the latter, used in sequence-to-sequence applications (e.g., question answering or machine translation), the decoder produces the output sequence.

In both configurations, the transformer works in two phases. In the  first phase, the transformer encodes the input into an intermediate representation $\mathbf{z}=(z_1,\ldots,z_n)$ where the dimension of the vector is fixed, typically where $d_{model}=768$. In the second phase, the transformer decodes $\mathbf{z}$ to produce the output. For example, in sequence-to-sequence generation the output would be a sequence of tokens $\mathbf{y}=(y_1,\ldots,y_l)$, ending with a special token. 

The model contains two stacks of repeating layers. The stacks differ slightly in composition: one stack is for encoding an input sequence, $\mathbf{x}$, to $\mathbf{z}$ and the other stack is for incrementally decoding the output $\mathbf{y}$ from $\mathbf{z}$, using the partially generated $\mathbf{y}$ as context when decoding. In each stack, the repeating layer contains a multi-head attention mechanism, that weighs the importance of tokens given the context it appears in, as well as fully connected network that independently performs a transformation on the attended tokens. A transformer model is composed of a fixed number of between $N=6$ and $N=12$ layers.

It is important to note the complexity of transformers. During encoding, self-attention considers the similarity between all tokens in the input sequence, which has quadratic complexity with respect to input sequence length. Similarly, during decoding, at step $t$, the attention mechanism scores the generated tokens $\mathbf{y}_{\lbrack 1: t-1\rbrack}$ against all of the encoded context $\mathbf{z}$, which is also quadratic. This complexity is clearly of concern when inputs are large.

\subsubsection*{Scaling NLP to DB scale}
The NLP problem of  question answering with external knowledge such as Wikipedia,  (a.k.a.\ {\em open-book QA}) forms a good starting point to explore the application of transformers to \ndb. In the context of \ndb s, we use  the transformer in  an encoder-decoder configuration, and the input contains the query to the \ndb\ and all the relevant facts in the database separated by a special delimiter symbol. The output is a sequence of tokens that answers the query.

To scale neural reasoning to databases of non-trivial size, it would not be feasible to encode the entire database as input to the transformer for a given query, because transformers cannot accept such large inputs, and even if they could, the latency would be prohibitive. It is common to use a maximum input size of $512$ or $1024$ tokens. The typical approach in open-book QA is to complement the transformer reasoning with an information retriever that extracts a small subset of the facts from the corpus. The information retrieval component can either be a simple one (e.g., BM25) or trained jointly with the transformer to learn to extract relevant parts of the corpus~\cite{Lewis2020,Guu2020,Thorne2018a,petroni2020kilt, Karpukhin2020, Wolfson2020}. However, the following additional challenges arise in the context of \ndb s: 

\begin{itemize}
    \item Unlike 
open-book QA, which typically requires extracting a span from a single document or predicting a token as an answer, answering queries in a \ndb\ may require processing a large number of facts and in some cases performing aggregations over large sets.

\item \ndb s do not enjoy the locality properties that usually hold in open-book QA. In \ndb s, a query may be dependent on multiple facts that can be anywhere in the database.  In fact, by definition, the current facts in a database can be reordered and the query answers should not change. In contrast, in 
open-book QA, the fact needed to answer a given question is typically located in a paragraph or document with multiple sentences about the same subject where this additional context may help information recall.

\item When determining which facts to input to the transformer, \ndb s may require conditional retrieval from the database. For example,
 to answer the query {\sf Whose spouse is a doctor?} we'd first need to fetch spouses and then their professions. In the NLP community this is known as multi-hop query answering~\cite{asai2019learning}, which has recently become an active area of research, but restricted to the case where we're looking for a single answer. In \ndb s, we may need to perform multi-hops for sets of facts.  

\end{itemize}

\section{Neural query processing}
\label{section:first-experiment}

In this section we describe an initial experiment whose goal is to better understand and quantify the 
applicability of transformers to query processing in \ndb s. In particular, the goal of the experiment is to answer the following question. Given a query and a small number of facts from the database, can the transformer accurately answer queries that are posed in natural language, whose answer may require projection (i.e., extracting part of a sentence), join and aggregation. Note that for the purpose of this experiment, we are momentarily putting aside the issue that transformers can only take a small number of facts as input. We'll address that issue with our full architecture in Section~\ref{section:architecture}.  

\begin{figure}
    \centering
    \includegraphics[width=\linewidth]{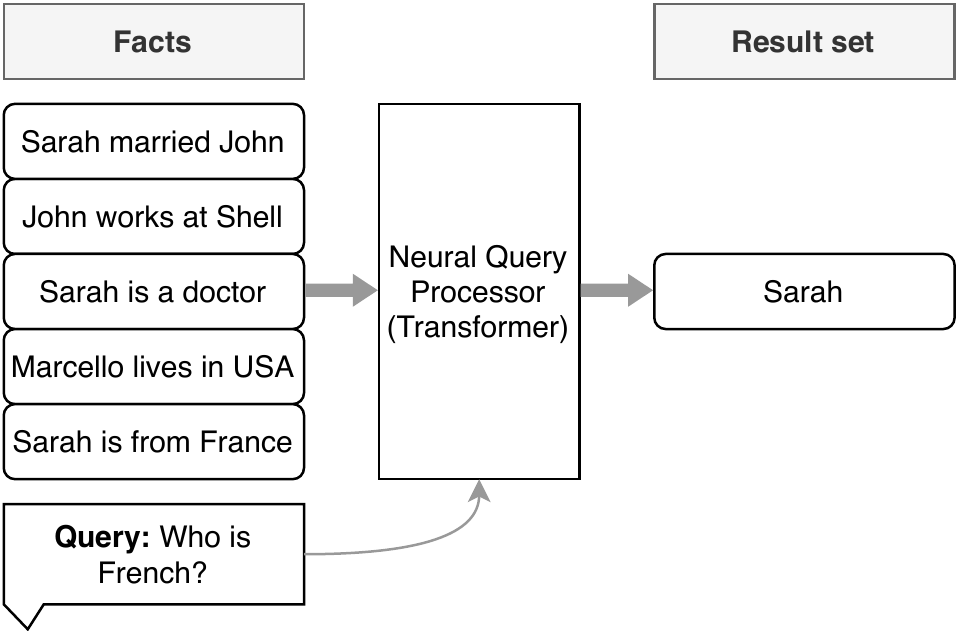}
    \caption{Prototype neural query processor using a T5 Transformer. The facts and the query are concatenated and given as input to the transformer. Our results show that transformers can answer lookup and join queries when given the support set of facts needed to generate an answer, but that the architecture does not scale to many facts.}
    \label{fig:nqp}
\end{figure}

\subsection{Data}
\label{section:datasets}

Since \ndb s are a new kind of database, there are no existing data sets that are directly applicable to evaluating them. Hence we now describe a data set we developed to test \ndb s, and to share with the community. While this dataset does not include {\em data in the wild}, it has enough variety that it provides a good signal about the validity of our techniques. 

Training a \ndb\ requires supervision in the form of $(D, Q, A)$ triples, where $D$ is a set of facts, $Q$ is a query and $A$ is the correct answer to $Q$ over $D$. We generate training data in a controlled fashion using data from Wikidata~\cite{vrandevcic2014wikidata} to express facts in natural language.
Because of the scale of Wikidata, it is possible to generate large numbers of training instances about a wide range of relationships requiring very few templates. 
The data set we create enables us to drill down our analysis by query type and relation type to understand the performance limitations of \ndb s. 
%\ah{The sentence below is hard to understand. Is it talking about out-of-domain data vs in domain? if so, let's make that clearer.}
%We show that new relations can be on-boarded to the \ndb\ after training using these data sets which would allow adaptation to other domains as training data becomes available. 

\subsubsection*{D1: Query and Answer dataset}

Wikidata stores triples of the form (S,R,O), where $R$ is a relationship between the subject $S$ and the object $O$, e.g., {\sf (Bezos, employedBy, Amazon)}. For every relationship we consider, we construct multiple natural language templates, thereby providing different linguistic forms of stating the same fact. Each template has a placeholder for the subject and the object, e.g., {\sf \$O is the place of birth of \$S}. We then generate different data instances by substituting entities from Wikidata for the placeholders in the templates.

We create databases with $7$ different relationships and for each relationship we create between $5$ and $14$ templates which vary pronouns, temporal expressions and phrasing of the fact and query.
We generate a training, validation and held-out test set containing $535$, $50$ and $50$ databases respectively, each containing $50$ facts. Each database has between $100$-$200$  question and answer pairs yielding $60000$ training, $5500$ validation and $6000$ test instances in total. 

The facts that are generated from Wikipedia are consistent with real-world facts. Hence, there is a risk that the \ndb\ is getting its answers from the pre-trained language model (trained on Wikipedia) and not by processing facts in the database itself. We mitigate this issue in two ways.
First, we create some facts about fictional characters with simple traits (e.g. {\sf likes coffee} or {\sf good at music}). For these facts, we use relationships and entities that are not in Wikidata (the entities are referenced by first name only). 
Second, we attach a timestamp with each query, and only facts prior to this timestamp are used for inference. By purposefully setting the timestamp lower than the timestamp of the relevant facts, we verify that the model is returning the {\tt NULL} answer when it's supposed to and not relying on facts in the pre-trained language model. %\ms{this is the first time we are mentioning NULL and we don't explain what NULL mean, should we add NULL to query types in the earlier section?}

%as well asuch as countries and geographical relations are consistent with the real world.
%To evaluate whether the model is memorizing all the relations and entities from Wikipedia during language model pre-training, or when fine-tuning on our databases. 
%If this occurs, the model wouldn't be learning to read facts from the database at query-time, instead relying only the parametric information that will not change after training is completed. Two features of the dataset account test for this: (1) each query is assigned a timestamp $t_i$ for which  each database,, where the fact is not present, the expected output is {\tt NULL}. 
%we added two additional relations for each person: (S,``employedAt'',O) where ``O'' is an organisation and (S,``is'',O) where ``O'' is a personality trait e.g. ``hard-working''. Personality trait does not exist as a relation in Wikidata. For the relation ``WorksAt'', we assign a random organisation to each person, where the organisation type may not match the profession of the person, e.g. {\sf Sandro is employed by Colt Telecom Group plc as a screenwriter.} 
%In this case, the model faced with the question {\sf Is Sandro  employed by Colt Telecom Group plc?} can not utilise its world knowledge that is stored in its parameters to reason about the answer, and has to memorise the facts from the database. 

The dataset contains both atomic and composite facts: composite facts combine more than one relation over the same subject, e.g., {\sf John {\tt [subject]} is employed as a driver {\tt [object(employedAs)]}\ in London {\tt [object(employedWhere)]}}.
To generate queries that require joins we apply the same technique to combine two connected relations for the query rather than the fact. For example, we use  ``fatherOf'' and ``employedAs'' to create the template for the query {\sf Does \$S's father work as a \$O}. To generate queries that require implicit language reasoning and test the knowledge captured by the language modelling task, we replace place names with a more general location. For example, for the fact: {\sf Mahesh's mum gave birth to him in Mumbai}, the questions generated would ask: {\sf Was Mahesh born in Europe?} (with the answer {\sf No}) and {\sf Was Mahesh born in India?} (with the answer {\sf Yes}).

%\jt{TODO LOW EASY: describe number of relations/properties expressed in D1}

\subsection{Results}
In our experiment, we use a T5 transformer model~\cite{Raffel2020}, a transformer variant that is designed for conditional language generation, as a neural query processor.  
To provide input to the transformer, we jointly encode relevant facts from the database by concatenating them with the query~(separated by a special delimiter token), as illustrated in Figure~\ref{fig:nqp}. In what follows, we consider several variants on which input facts are encoded by the transformer which both provides an upper bound on performance (if all necessary facts were encoded), as well as evaluates the transformer's resilience to noisy (low precision) or incomplete (low recall) information retrieval when performing neural query processing.  

\begin{figure}[h]
    \centering
    \includegraphics[width=\linewidth]{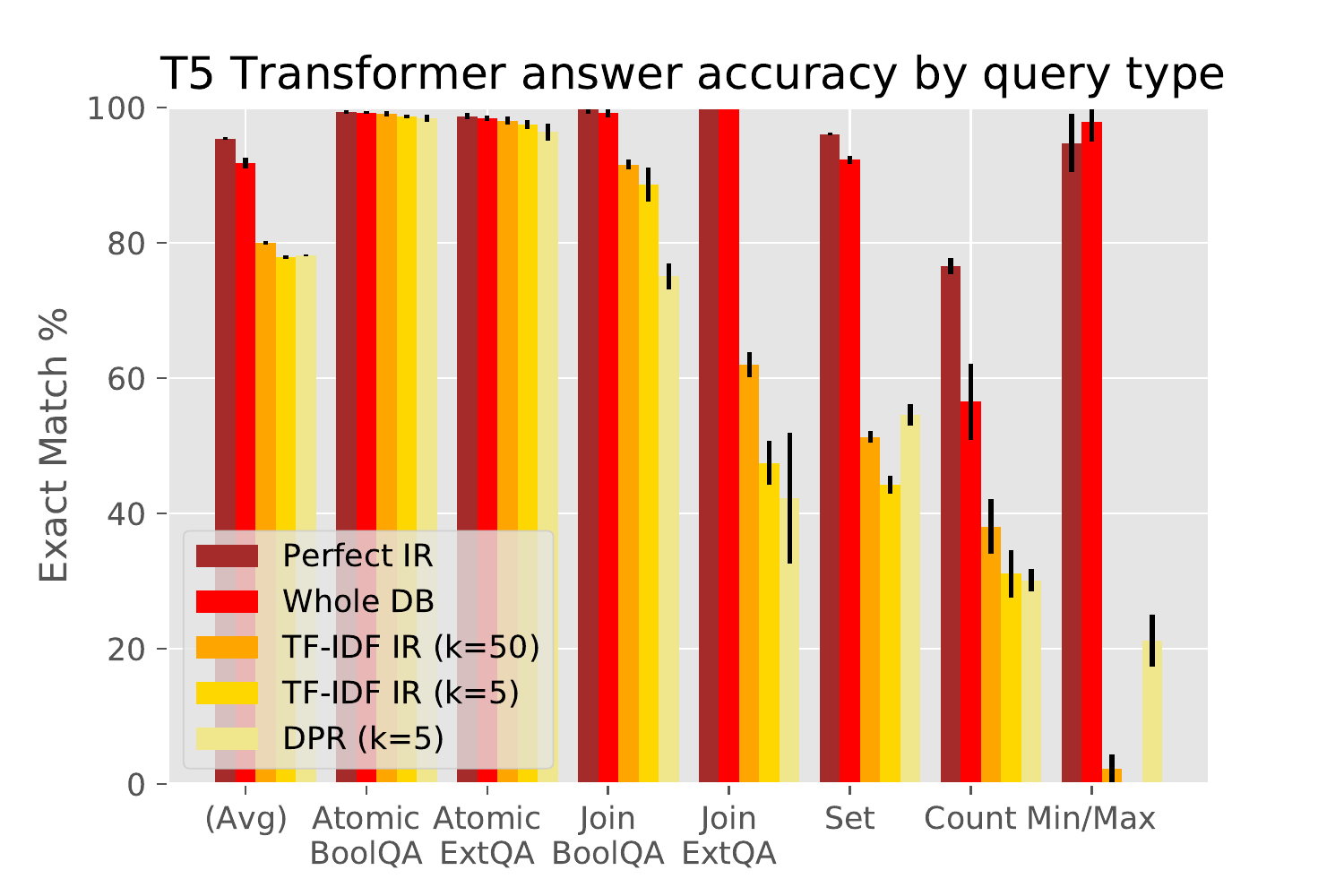}
    \caption{Exact match accuracy for different classes of queries. The results show that the transformer obtains high accuracy for lookup and join queries, but falls short for queries with aggregation or yielding set answers. Furthermore, adding information retrieval to scale to larger databases harms EM for queries outputting a set of results or requiring aggregations.}
    
    \label{fig:neural-everything}
\end{figure}

%We perform this evaluation on a collection of reference databases containing no more than $50$ updates. 

\subsubsection*{Varying the inputs to the transformer}
We first investigate how resilient the transformer is to the number and relevance of its input facts. Figure~\ref{fig:neural-everything} shows the exact match scores for several ways of retrieving/filtering facts before they enter the transformer, with respect to different query types. Examples of correctly answered queries for the Perfect IR model are shown in Figure~\ref{fig:actual_data}.

Our first observation is that for queries which required either extracting information from the facts, or performing Boolean inference, the model attained near perfect scores, regardless of whether the queries need to be answered from a single fact or by joining multiple facts. The fact that the model had high scores for queries that require the combination of multiple facts indicates that the transformer model is able to combine information from multiple sources when generating an answer to the user query. The different variants of our experiment provide further insights. 

\medskip
\noindent\textbf{Perfect IR:}  This approach assumes that the information retrieval component of model can perfectly retrieve the set of facts needed to answer a query.  This version assesses the model's capability of performing the right computations when only the appropriate facts are given as input to the model. To implement it we select the appropriate facts using meta-data from construction of the controlled reference dataset.

Our results suggest that given the right facts, the model can be robust to multiple linguistic variations and generates the correct output. However,  the model performs poorly for query types where an aggregation step is necessary or when the query result is a large set. 
This observation corroborates work by Hupkes \emph{et al.}~\cite{Hupkes2020} that showed that neural models can't generate long sequences well for certain sequence transduction tasks.

\medskip
\noindent\textbf{Whole DB:} in this the approach we encode the whole database as input to the model. This would only work for small databases (it is prohibitive even for databases with 50 facts) because the self-attention mechanism in the transformer model has quadratic complexity\footnote{While optimizations and linearizations have recently been introduced \cite{Wang2020,Stickland2019} a single encoder would present a bottleneck. Furthermore aggregating large number of facts (such as with {\tt count}) would still suffer from low accuracy.} with respect to input size. This method allows us to evaluate the model's sensitivity to noise and whether having too much information (i.e. low precision with high recall) has an impact. 

%This approach has the highest overall EM score. 
One positive takeaway is that the model is able to generate the correct answers despite the fact that it was exposed to irrelevant facts. Interestingly, this model also has a high $F_1$ score for queries that generated sets as results (as opposed to a single fact) as well as for min/max queries. For count queries its performance is low.

\medskip
\noindent\textbf{\tfidf\ IR:} In this version we evaluate a simple baseline for the information retrieval component using \tfidf\ which considers overlapping tokens between the query and fact. This evaluation would highlight whether the \ndb \ model adequately handles noise from the IR component as well as the difficulty of retrieving the necessary facts. We may fail to retrieve the relevant facts either due to the need for multi-hop reasoning or because facts are expressed with different linguistic expressions. Our experiment considers using both the top-5 and top-50\footnote{The DB size is 50 facts. This evaluates whether the limitation in \tfidf\ is the requirement for token overlap or whether the limitation is in ranking} returned results from the \tfidf\ component, and evaluating whether a small number of top results is sufficient for some query types. 

Using \tfidf\ for IR, the model still attains near-perfect accuracy for atomic queries.
For queries where a join is required, the accuracy drops due to low recall. While the EM from Boolean QA is reduced from 99.2\% (using the whole DB) to 91.6\%, for queries that require extracting information the accuracy drop is more stark: from near perfect to 62.0\%. From a modeling perspective, Boolean query answering is quite a simple simple task with low entropy that the model may be able to guess using other cues in the query or facts (i.e., it may be correct for the wrong reasons). The accuracy for queries that output sets, or require aggregation is much lower. 
For queries with min/max aggregation, the EM for \tfidf\ was near zero, this is due to there being no token overlap between the query (such as {\sf Who is the oldest?}) and the facts (such as {\sf John was born in 1962}), yielding no results from the \tfidf\ search engine.

\medskip
\noindent\textbf{DPR:} To alleviate some of the limitations of \tfidf, we experimented with  dense passage retriever (DPR)~\cite{Karpukhin2020}, which scores the similarity of vector encodings of the query and the facts through computing the inner product assigned a non-zero score for all fact (meaning that all relevant facts could potentially be retrieved - even if there are no overlapping tokens).
%With a sufficiently large context size, any fact could be input into the model when using DPR for information retrieval.
Our experiments, however, highlight a difficult precision-recall trade off. To retrieve facts for joins, many false positive facts must also be input into the model as we find that facts participating in joins are often ranked outside of the top $5$ items returned from the information retrieval.

\begin{figure}[]
    \centering
    \includegraphics[width=\linewidth]{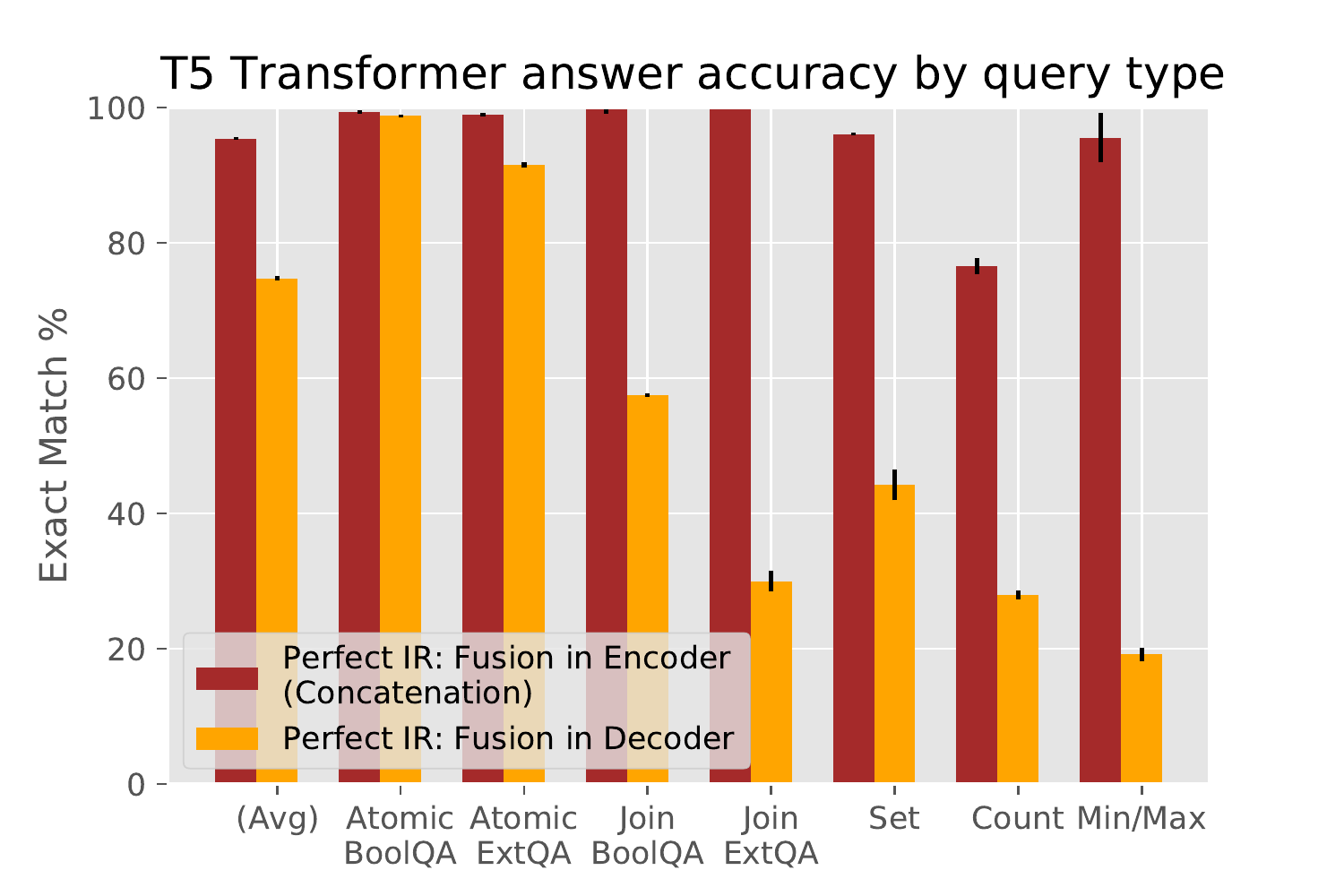}
    \caption{Fusing in the decoder, as opposed to concatenating the inputs (hence fusing in the encoder), reduces the transformer's computational complexity, but accuracy drops significantly. This experiment suggests that \ndbtitle s should only process small numbers of facts in every transformer while fusing facts in the encoder. }
    
    \label{fig:fid}
\end{figure}

\subsubsection*{Independent encoding of facts with Fusion in Decoder} 
The following experiment provides an additional insight that is useful for developing the architecture for \ndb s. In the experiments so far, we concatenated all the facts before we fed it to the encoder of the transformer. All facts were encoded jointly with self-attention which both considers intra- and inter-fact relations.
Here we adapt the approach known as Fusion in Decoder (FiD)~\cite{Izacard2020} to our context. Specifically, each fact is fed separately to the encoder, but the fusion of the facts happens only in the decoder. Rather than one large self-attention operation over all facts, self-attention is computed independently for each fact, considering the relation between the fact and the query, but there is no attention between facts. This reduces the complexity of joint encoding of multiple concatenated facts from a single quadratic complexity operation (w.r.t.\ total input length) to a linear (w.r.t.\ number of facts) allowing this method to scale to larger numbers of facts.

We compare FiD with the aforementioned \emph{perfect-IR} version, where the only input to the models are the facts necessary to answer the query. The results in Figure~\ref{fig:fid} show that while the two models perform comparably for lookup queries,  the FiD approach fails for queries that require join or aggregation. 
The implication of this experiment is that the self-attention in encoding is important to capture the inter-sentence dependencies between the facts. Therefore, trying to feed growing numbers of facts to a transformer is unlikely to scale.

\subsubsection*{Summary}

We believe that the initial experiment suggests the following: (1) if there were a way to feed the transformer the relevant facts from the database, it can produce results with reasonable accuracy, (2) aggregation queries need to be performed outside of the neural machinery, and (3) in order to handle queries that result in sets of answers and in order to prepare sets for subsequent aggregation operators, we need to develop a neural operator that can process individual (or small sets of) facts in isolation and whose results outputted as the answer or fed into a traditional (i.e. non-neural) aggregation operator. The next section describes our first steps towards such an architecture.

\jt{TODO LOW MEDIUM: list recall scores- how well does the IR work
* The recall for the \tfidf\ model is not perfect. This doesn't affect the model's ability to reason with missing information?
Accuracy for BoolQA over a join is high for the \tfidf\ model - model is memorizing some facts without needing all the evidence. EM drops with Join EXtQA.}

\jt{TODO LOW EASY: for set questions, the model hallucinates things multiple times. even with perfect IR}

\section{The architecture of a Neural Database}
\label{section:architecture}

\begin{figure*}[ht]
    \centering
    \includegraphics[width=\linewidth]{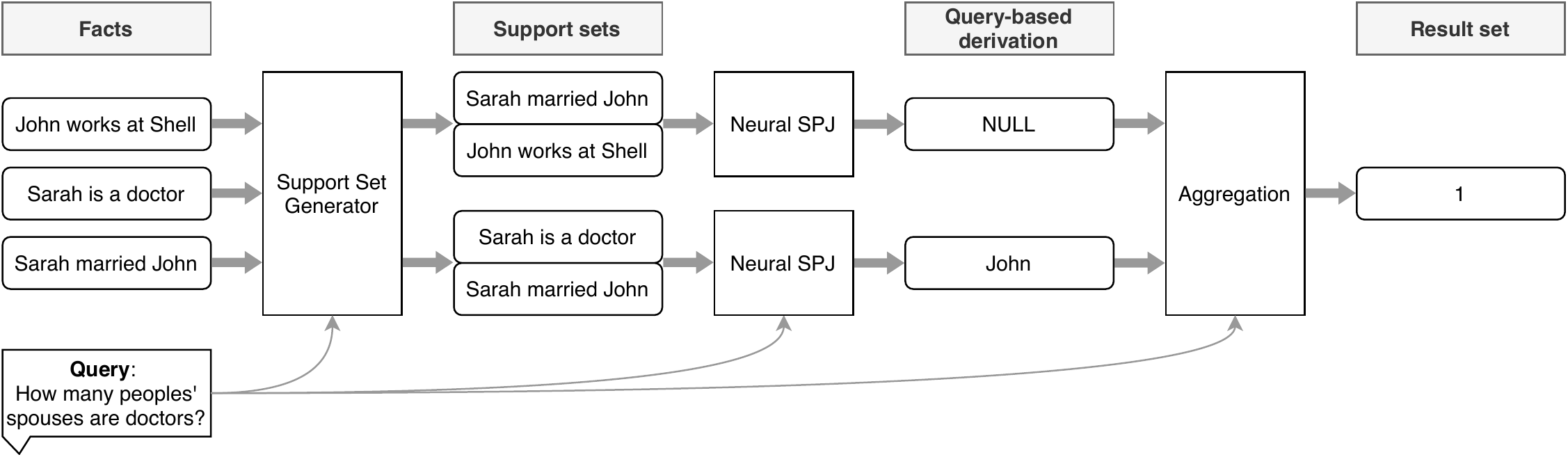}
    \caption{Overview of \systemname\ architecture. The support set generator creates small sets of facts that are each fed into a separate Neural SPJ operator that runs a single transformer. The results of the individual Neural SPJ operators are either unioned to produce the result or passed on to a traditional aggregation operator.}
    \label{fig:overview}
\end{figure*}

We now describe the architecture of a \ndb\ that addresses the challenges exposed in Section~\ref{section:first-experiment}.
The architecture is shown in Figure~\ref{fig:overview} and has the following key ideas:

\smallskip
\noindent
{\bf Running multiple transformers in parallel:}
In Section~\ref{section:first-experiment}, we demonstrated that if we provide our transformer model with the right facts needed to derive the query, and even in the presence of irrelevant facts, the transformer can produce correct answers for SPJ queries. The problem is that we can only feed a small number of facts to the transformer.

In our architecture we address this challenge by running multiple copies of a Neural SPJ operators in parallel. Each copy is a transformer similar to the one we used in Section~\ref{section:first-experiment}.
When queries don't involve aggregation, the union of the outputs of the Neural SPJ operators are the answer to the query. When the query does involve aggregation, these machine-readable outputs are fed into the aggregation operator.  

\smallskip
\noindent
{\bf Aggregation with a traditional operator:} Since the Neural SPJ operators were designed to output structured results, our  architecture can use a separate traditional aggregation operator.
Using a separate aggregation overcomes the limitation on transformers demonstrated in Section~\ref{section:first-experiment}, and enables us to extend the system to new aggregation operators without retraining the basic models used in the system. \fs{In my opinion the previous sentence is not well supported. What does it mean to extend with new aggregation operators?}
% A query analysis component determines whether the query requires an aggregation function, and if it does, which aggregation is needed for the query (e.g., count or max). \ah{Two sentences here on how this component is trained.}
The aggregation operator is selected through a classifier we train separately that maps the query to one of 6 labels: \{no\_aggregation, count, min, max, argmin, argmax\}.

\smallskip
\noindent
{\bf Support set generation:}
%The supports sets are generated by the Support Set Generator~(SSG), another novel component of the architecture.
Intuitively, the answer to every SPJ query has a derivation tree, where the leaves are the facts from the database. The Neural SPJ operator needs to be given a preferably small superset\footnote{In a perfect model, the SPJ operator should be given exactly the leaves of the derivation tree, but this is challenging. We optimize the SSG generator for recall and the SPJ for precision over noisier support sets} of these leaves. More formally, 
each copy of the Neural SPJ operator is given a \emph{support set}: a restricted subset $\hat{D} \in \mathcal{P}(D)$ of the facts in the database that contains the minimal \fs{the minimal reads as it is the smallest possible but I take it we are inputting a small-ish support set. Am I wrong?} support to generate an answer to the query.

\subsubsection*{Training the Neural SPJ operator}
\begin{figure}[h!]
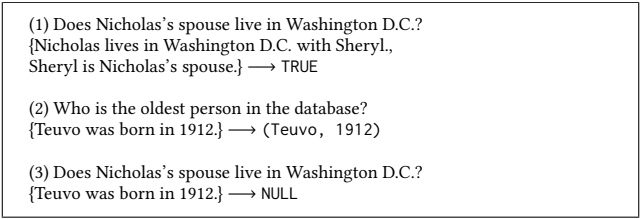

    \centering
\noindent
\footnotesize
\begin{mdframed}
\begin{tabbing}
(1) Does Nicholas's spouse live in Washington D.C.?\\
\{Nicholas lives in Washington D.C. with Sheryl.,  \\Sheryl is Nicholas's spouse.\}
$\longrightarrow$ {\tt TRUE} \\ \\

(2) Who is the oldest person in the database? \\
\{Teuvo was born in 1912.\} $\longrightarrow$ {\tt (Teuvo, 1912)}\\ \\

(3) Does Nicholas's spouse live in Washington D.C.?\\
\{Teuvo was born in 1912.\} $\longrightarrow$ {\tt NULL}
\end{tabbing}
\end{mdframed}
    \caption{Examples of the intermediate results that are produced by the Neural SPJ operator. }
    \label{fig:query_derivations}
\end{figure}

The Neural SPJ module is a transformer model. Unlike the prototype model described in Section~\ref{section:first-experiment}, which was trained to generate the final answers to the query, the Neural SPJ is trained to generate an intermediate result of the query. Figure~\ref{fig:query_derivations} shows examples of the output of the Neural SPJ operator.

% In Section~\ref{section:experiments} we describe how training data for this task can be generated at scale through templates on Wikidata, demonstrating success for in-domain reasoning (for relations captured in training data) and out-of-domain reasoning (for new relations that the model has not been trained on). 

%We also demonstrate how question-answer pairs over entire database (used to train the prototype model in Section~\ref{section:first-experiment}) can be adapted to train this model.

% The Neural NPJ module works under the assumption that, given the necessary facts as input, a neural model can generate an appropriate derivation\footnote{The discussion of generating appropriate support sets with the SSG module is left to Section~\ref{section:architecture:support-set-generation}}. 

% The derivation generated depends both on the query and the facts, performing an implicit join if needed, shown in Figure~\ref{fig:query_derivations}. For aggregations, the query-based derivation is a machine-readable tuple containing an entity and, if applicable, a numerical value. For Boolean question or extractive question answering over the database, the query-based derivation is the answer to the query. For candidate sets that do not contain facts required to answer the query, the query-based derivation is a {\tt NULL} value, filtered prior to aggregation.

\section{Support set generation}
\label{section:architecture:support-set-generation}

 The Support set generator (SSG) is a module that given a query $Q$ and a database $D$, produces a set of support sets $SSG_Q(D)$, each of which is used to generate an answer with the SPJ module in parallel.
%Each support set $S \in SSG_Q(D)$ is a set of facts in the database that could produce an answer to the SPJ component of the query Q. 
Note that sets in $SSG_Q(D)$ may not be pairwise disjoint because some facts may be required for multiple answers (consider, for example, a one-to-many relation). 

The outputs generated by SSG depend on the information need of the downstream SPJ, and whether the query requires joins or aggregation: 
(1) for queries that are answered by a single sentence, e.g., {\sf Who is Sheryl's husband?}, the support set containing a single fact should be generated, e.g.,   {\sf Sheryl is Nicholas's spouse}. 
(2) When the SPJ operator requires multiple facts to be joined, the support set would contain all participating facts. 
(3) For queries that require aggregation or whose answer is a set, multiple support sets must be generated, each containing enough information to generate the intermediate results that are aggregated.
For example, for the query {\sf Who is the oldest person?}, each of the support sets would contain a single fact that includes a person and their birth date.  If joins are required to generate the intermediate result, the support sets would contain multiple facts.

Using information retrieval, such as \tfidf\ in Section~\ref{section:first-experiment}, could be considered a primitive SSG generating a single support set containing all relevant facts.
As our experiments indicated, this method worked well for queries whose answer is generated from a single fact, but not for joins, aggregation queries or   for queries outputting a set of answers. 
%for two reasons.
In what follows we describe a more robust algorithm for generating support sets.

% For queries requiring joins, multi-hop retrieval is needed, where the intermediate set of facts found dictate how the information retrieval module should find more information. And for queries requiring aggregations or outputting a set of results, the single set of facts returned by the information retrieval would need separating into multiple support sets to allow multiple Neural SPJ operators to independently generate the answers.

%for the query {\sf How many people's spouses are doctors?}, we may retrieve the fact {\sf Sarah is a doctor}, but only then would we try to retrieve facts about Sarah's spouse.
%\MY{additionally IR can not make a set of sets for aggregation stuff, perhaps we need another module to find the support set seeds, then run IR module for each seed}

\subsubsection*{Incremental support set generation}

\begin{figure}[ht]
    \centering
    \includegraphics[width=\linewidth]{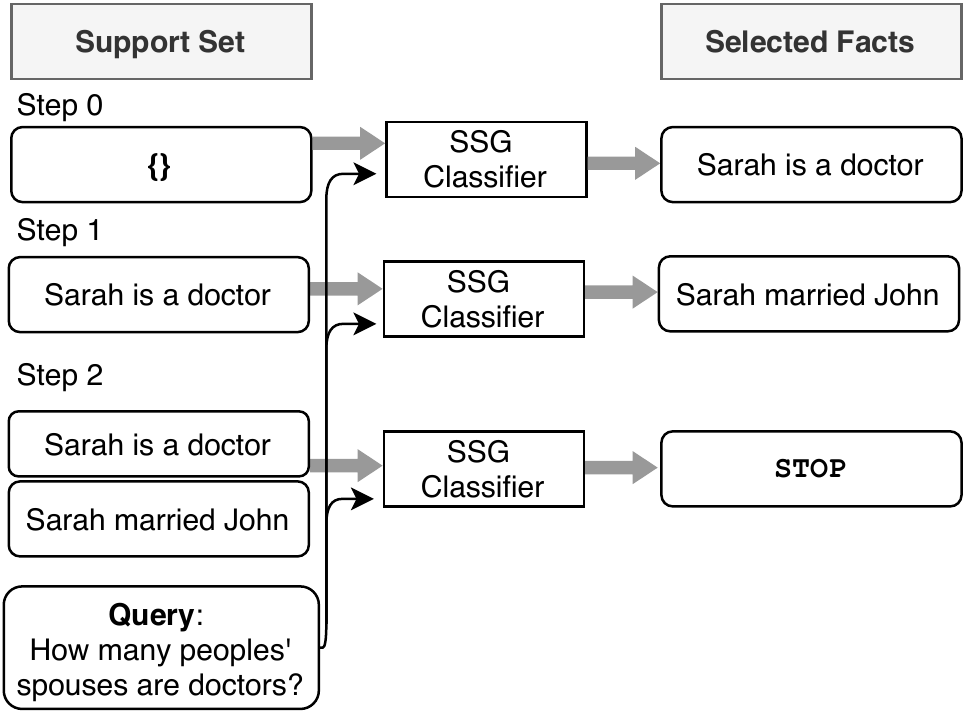}
    
    \caption{ISSG incrementally creates support sets. At each step, the classifier either decides to add another fact to the support set or to stop and output a completed support set.}
    \label{fig:issg-illustration}
\end{figure}

\begin{algorithm}[t]
 \SetKw{In}{in}
 \KwIn{Bi-encoders $C$: $C_U$ (for actions), $C_V$ (for state), Database $D$, Query $Q$, Threshold $\tau$}
 \KwOut{Set of support sets ($\hat{D}_1,\ldots,\hat{D}_b$) $\subset \mathcal{P}(D)$ }
 %$M$ is the binary classifier that receives a state and an action, and decides whether to take that action or not 
 open := \{\{\}\}\; 
 closed := \{\}\; 
 U := $\lbrack C_U(u_1); \ldots; C_U(u_n); C_U(\mathtt{STOP}) \rbrack$ for $u_i \in D$\;
 
 \While{open $\neq$ \{\}}{
  next := \{\}\;
  
  \For{$\hat{D}_k$ \In open}{ 
    V := $\lbrack C_V(Q, u_1 \ldots u_m)\rbrack$, for $u_i \in \hat{D}_k$\;
    
    A := MIPS($U$,$V$,$\tau$)\;
    \For {$a_j$ \In $A$} {
        \If{$a_j$ == \texttt{STOP}} {
            $closed := closed$ $\cup \{\hat{D}_k\}$;
        } 
        \Else{
            $next := next$ $\cup \{ \{ a_j \cup \hat{D}_k\}\}$\;
        }
        
    }
          
    }
    
  open := next\;
 }
 \Return closed;
 \caption{Support Set Generator (SSG) modeled as multi-label classification: using maximum inner product search (MIPS) over vector encodings of facts $U$ and state $V$ }
 \label{ssg-alg}
 %\vspace{-1em}
\end{algorithm}

The output of the SSG is the set of all relevant support sets: $SSG_Q(D) \subset \mathcal{P}(D)$. It would be intractable to consider all possible support sets, as this would be akin to enumerating the powerset. We instead efficiently and incrementally construct support sets, starting from the empty set by modeling the task as multi-label structured prediction. 
At each step, a classifier, $C$, considers the partially generated support set $\hat{D}_k$and the query and predicts which candidate facts $u_i \in D$ from the database should be added or whether to stop the iteration.

Incremental SSG is described in Algorithm~\ref{ssg-alg} and illustrated in Figure~\ref{fig:issg-illustration}. The action classifier predicts which facts $u_i \in D$ should be explored or whether to stop. If {\tt STOP} is predicted, $\hat{D}_k$ is closed (i.e., it forms part of the output); otherwise, for each fact added, a new intermediate (i.e., open) support set is generated which is explored in the next iteration.
For efficiency, to build the multi-label SSG action classifier, we use a bi-encoder architecture that independently encodes the facts in the database and the state (query and a partial support set) and computes the inner product between the encoded representations to generate a score. The encoders are pre-trained transformers fine-tuned to yield a high inner product between the state's encodings and relevant facts to be added to answer the query. The vectors encoding of the facts are static and are pre-computed offline.
At each step, $t$, we encode the state using a transformer by concatenating the query and all facts already included in $D_k \in open $. 

% The details of the encoder's pre-trained transformer is given in section~\ref{section:experiments:ssg}.

\smallskip
\noindent
{\em Complexity:} The inner loop of Algorithm~\ref{ssg-alg} involves a Maximum Inner Product Search (MIPS) between the encoded state and the encodings of the facts, which linear in the number of facts. However, we use FAISS~\cite{Johnson2019} to accelerate retrieval to $O(\log^2 n)$. If we assume a query needs a maximum of $b$ support sets, and the average size of a support set is $m$, then the complexity of the SSG algorithm is
$O(bm\log^2 n)$. Both $b$ and $m$ are bounded by the number of facts in the database $n$, but in practice we'd expect only one of $b$ or $m$ factors to be large. However, there is fertile ground for developing methods for indexing (and/or clustering) the facts in the database so that only few facts need to be considered in each iteration of the inner loop of the algorithm, leading to significant speedups. 

\begin{algorithm}[t]
 %\SetAlgoLine
 \KwIn{$Q$ the NeuralDB for question, $D$ all facts} 
 \KwIn{$maxIters$ optional control for early stopping} 
 \KwResult{Support set $\hat{D} \subseteq D$ such that  $Q(D)= Q(\hat{D})$ }

 \SetKw{Yield}{yield}
 \SetKw{In}{in}
 \SetKw{YieldFrom}{yield from}
 \SetKwFunction{FPredictPath}{Predict}
 \SetKwProg{Fn}{Generator:}{ is}{end}
 
 \Fn{\FPredictPath{D, reference, history}}{
    \If{$D$ -- history = \{\}} {
        \Return \{\};
    }
    
    search = random.shuffle($D$ -- history); \\
    \For{fact \In search}{
        predicted = Q(D--\{history $\cup$ \{fact\}\} ); \\
        \If{predicted $\neq$ reference} {
            \Yield {fact} \\
            \YieldFrom{Predict(D, reference, history)}
        } 
        history := history $\cup$ \{fact\}; 
    }
 }
 i := 0; $\hat{D}$ := \{\}; ref := $Q(D)$; \\
 \While{fact := Predict(D, ref, \{\}) $\wedge i < maxIters$}{
    $\hat{D} := \hat{D} \cup \{fact\}$;  \\
    i += 1;
 }
 return $\hat{D}$\;
 \caption{Generating training signal for distant supervision of support sets by querying a NDB with perturbed facts which the triggers the generated answer to change}
 \label{alg:dfs}
\end{algorithm}

\subsubsection*{Training the action classifier} 
To train the action classifier, we need a supervision signal as to which facts the classifier must select at a given the current state for a query. We generate such training data from the data set $D1$, where we have a set of queries, answers, and their corresponding support sets.

% In section~\ref{section:experiments:ssg} we show the performance of the supervised SSG experimentally.

To alleviate the need for training data we describe a novel distant supervision approach for training the action classifier (see Algorithm~\ref{alg:dfs}). 
Instead of having perfect training data, we generate possibly noisy training data by taking advantage of the neural SPJ operator itself. Specifically, with a known (query, answer) pair and a pre-trained model, we can incrementally remove facts from the database until the answer predicted by the model changes from the correct answer to one which is incorrect. We know that for small databases, it is possible to encode the entire database as input to the SPJ operator, so we use the whole DB model we pre-trained from Section~\ref{section:first-experiment} for this purpose.
The combination of facts removed from the DB that change the answer from the correct answer to an incorrect answer would be labeled as forming part of a support set. 
For example, removing the either the fact {\sf Sarah is a doctor} or {\sf Sarah married John} from the input to the model may change the output prediction for the {\tt How many people's spouses are doctors?} query from $1$ to $0$, and would be added to the support set.
The training data may be noisy because the neural SPJ operator is not perfect: robustness to stochasticity could be introduced through reinforcement learning techniques such as Q-learning~\cite{Watkins1992}, but investigating that option is reserved for future work. 
Experimental results in Section~\ref{section:experiments:ssg} indicate that the models can be robust to this noise without explicitly mitigating the noise. Our 
 method is a form of erasure-based black-box model explanation technique~\cite{Ribeiro2016, Li2016}. However, we treat facts as features rather than individual tokens.

\section{Experiments}

\begin{table}[]
\caption{Exact match scores showing our proposed \ndbtitle\ architecture of SSG, SPJ and Aggregation accurately answers aggregation and join queries whereas IR(k=5) with a single T5 transformer does not.}
\label{table:pipeline}
\begin{tabular}{lccccc}
\toprule
\multicolumn{1}{c}{\multirow{2}{*}{\textbf{Method}}} & \multicolumn{5}{c}{\textbf{Exact Match (\%)}}                                                  \\
\multicolumn{1}{c}{}                                 & \textbf{Count} & \textbf{Min/Max} & \textbf{Sets}  & \textbf{Atomic} & \textbf{Joins} \\ \midrule
\ndb                                 & \textbf{79.31} & \textbf{100.00}  & \textbf{94.56} & \textbf{98.72}  & \textbf{96.95} \\
\ndb\ (DS)                              & \textbf{79.45} & \textbf{100.00}  & \textbf{91.91} & \textbf{97.90}  & \textbf{79.29} \\ \midrule
\tfidf +T5                                         & 31.06          & 0.00             & 44.25          & 98.05           & 68.02          \\
DPR+T5                                            & 38.07          & 21.19            & 54.55          & 97.38           & 58.64          \\ \bottomrule
\end{tabular}
\end{table}

\label{section:experiments}
We begin in Section~\ref{section:experiments:full} by demonstrating the accuracy of the end-to-end \ndb, showing  that queries can be answered with high accuracy over thousands of facts.  We then validate the components of our architecture in isolation. 
In Section~\ref{section:experiments-architecture} we consider the basic architecture, consisting of Neural SPJ followed by aggregation (without the SSG). In section~\ref{section:experiments:ssg} we evaluate our SSG  module.
In Section~\ref{section:experiment-analysis} we discuss how the results depend on the amount of training data and how learning transfers to relations unseen in the training set. 

%In this section we set out to validate the components of our architecture. In Section~\ref{section:experiments-architecture} we demonstrate the basic architecture, consisting of Neural SPJ followed by aggregation, yields results with high accuracy. In section~\ref{section:experiments:ssg} we evaluate our SSG  module, and in Section~\ref{section:experiments:full} we evaluate the accuracy of query answering when the SSG provides the support sets. 
%In Section~\ref{section:experiment-analysis} we discuss how the results depend on the amount of training data and  how learning  transfers to relations unseen in the training set. 

\subsubsection*{Implementation}
We use the HuggingFace \cite{wolf2020huggingfaces} transformers library and its implementation of the {\tt t5-base} transformer module~\cite{Raffel2020} for SPJ. With the exception of one experiment, the parameters of the {\tt t5-base} model have been pre-trained on the colossal Common Crawl (C4) dataset~\cite{Raffel2020} by predicting tokens masked in sentences. %The {\tt t5-base} transformer comprises 12 layers, with the hidden units, feed-forward network and attention dimensions being 768, 3072, and 64, respectively. Each transformer layer has 12 attention heads. The total number of model parameters is 222 million. 
For SSG, we use {\tt BERT} which has a comparable architecture to T5. % and similar number of parameters. For aggregation, we use Python built-in methods.
The learning-rate for fine-tuning and number of epochs were selected through maximizing the Exact-Match (EM) accuracy on a held-out validation set for the tasks. Fine-tuning is a stochastic process: for each experiment, we independently train 5 separate models with different random seeds and report mean accuracy.

\subsection{End-to-end \ndbtitle}
\label{section:experiments:full}

Before validating the SSG and SPJ components independently, we first evaluate the full \ndb\ architecture end to end, reporting results in Table~\ref{table:pipeline}. At run time, the support sets generated by the SSG are input in to SPJ operators in parallel and then aggregated. Compared to state of the art transformer models (bottom 2 rows) with information retrieval, the \ndb\ achieves the highest EM accuracy on all types of queries. While these numbers cannot be directly compared (as the \ndb\ requires supervision to the intermediate results), the \ndb\ makes a substantial improvements over several query types, overcoming fundamental limitations in how the IR+transformer models perform aggregations. 
% The SSG operator and Neural SPJ have been designed specifically to address the shortcoming of the join retriever and T5 models. 
% The end to end \ndb Our \ndb\ also shows at least 11\% improvement on join queries too because of how the SSG operator iteratively adds related facts to the support set based on the query. These related facts are harder to be retrieved using \tfidf or DPR. 
It is expected that distantly supervising the SSG would decrease accuracy somewhat. %However, it had negligible impact for some query types and slightly lower performance over other query types as expected. 
However, given that the supervision signal is generated for free, this is encouraging, especially as the impact was negligible in some cases. The decrease in accuracy is most evident for joins because of the over sensitivity of the model used to distantly supervise the SSG, reflected in the low SSG recall in Table~\ref{table:ssg}.

% \ah{The results in figures 9 and 10 should be communicated in a different way. Right now the figures convey very little information compared to how much space they're taking.}

% \ah{The stuff below is kept for reference until this section settles.}
% The structure of the experiments section is as follows:

% \jt{
% 1) we demonstrate the neural SPJ operator can do selection, projection and join using a combination of data from D1+D2. How would neural SPJ affect downstream accuracy?
% 2) we look at ssg with and without supervision and how well it generates support sets
% 3) how well does Neural SPJ work with the output of SSG?

% 4) tying it all together (end2end) 
% 5) we look more into the pragmatics of training intermediate results (D2 only)
% 5a) does LM help relation transfer?
% 5b) how much data is needed?
% 5c) resilience to distractors/distribution shift over time?
% 5d) error analysis
% }

\subsection{Neural SPJ + Aggregation}
\label{section:experiments-architecture}

%\ah{I'm assuming this section shows the  performance of our proposed architecture. We run the SPJ then aggregation. The difference between here and section 3 is that we train the SPJ to output the intermediate result. This is probably the place to describe D2.
%}
The experiments in Section~\ref{section:first-experiment} showed that the transformer model was capable of generating the correct answer to queries requiring lookups and joins with near perfect accuracy on our test set when provided with the right facts or small supersets thereof. 
We evaluate the Neural SPJ on two settings. In {\bf Perfect IR}, we provide the Neural SPJ only the needed facts. Hence, Neural SPJ does not need to decide which facts to use to answer the query. In {\bf Noisy IR}, we sample random facts and add them to the support set used to train the model. Hence, we're also training the Neural SPJ to select the right set of facts to use for inference, thereby validating the resilience of the SPJ operator to extraneous facts that might be returned by a low-precision high-recall SSG. The noise was generated by sampling $1-3$ additional facts with uniform probability for $\frac{2}{3}$ of instances.

%Unlike the transformer in Section~\ref{section:first-experiment}, the Neural SPJ generates an intermediate result that, if needed, is aggregated by a downstream component. In this section we evaluate the Neural SPJ component of our architecture.

\subsubsection*{Dataset D2}
Training and evaluating the Neural SPJ operator requires data that differs slightly from the data in D1. In D1, the training data for a given query $Q$ involved the answer, $Q(D)$. However, for the Neural SPJ training we need to provide the result of the SPJ component before the aggregation. In D1, the output to the question {\sf How many countries use Euro?} would just be the number of countries, after aggregation. However, in D2, the training data  consists of a pairs of facts and intermediate results. For the fact, {\sf Belgium's currency is the Euro}, the intermediate result is {\tt Belgium} because the count operator needs the names of the countries that use the Euro. For the query {\sf What is the most widely-used currency?}, the intermediate result for the same fact  would include the pair ({\tt (Euro, Belgium)}) because the aggregation operator first needs to group by currency before computing the max. 

D2 includes queries for training the model to answer set and aggregation queries. We create a total of 632 templates: 115 for facts and 517 for the different query types over 27 relations, covering a a variety of linguistic expressions. Using a sample of popular Wikidata entities, we generate a single database of 8400 facts and 14000 queries over it.
For training, we generate approximately 10,000 training instances per aggregation operation, preventing class imbalance by preserving the proportions of D1.

\begin{table}
\caption{Accuracy of Neural SPJ + Aggregation portion of architecture. We use 200 parallel SPJ operators (4xV100 GPUs, batch size = 50) over a database with 8400 facts, averaged over 14000 queries.}
\label{table:spj}

\begin{tabular}{@{}lcc@{}}
\toprule
\multicolumn{1}{c}{\textbf{Input at Training}} & \textbf{Projection EM (\%)} & \textbf{Answer EM (\%)} \\ \midrule
Perfect IR                         & $97.87 \pm 0.1$          & $98.60 \pm 0.1$                \\
+ ({\em noise added at test})            & $76.25 \pm 2$            & $70.35 \pm 3$                  \\
Noisy IR                           & $98.21 \pm 0.1$          & $98.52 \pm 0.1$               \\

\bottomrule
\end{tabular}
\end{table}

%We also generate negative instances, one for every positive instance, that describe how the model should behave if irrelevant information is present in the support set. These are generated by using a query template and replacing the subject and object placeholders with entities such that the combination does not exist as a fact in the database. 
%For instance, the following is a negative instance, for the question {\sf Is Euro the currency of the USA?} over the fact {\sf Belgium's currency is Euro}, the expected derivation is {\tt NULL}.

% is the projection template would be the subject entity that is counted by a set cardinality function in the aggregation module: ``\$O''  and the output to the question  the output is a tuple used for a count-max by the aggregation module: .

%\ah{Do we need this here?}
%\jt{It might be useful for the top of the section - it's a nice property to have if it's also in D1 because it indicates the model can generalize to unseen entities} 
%The training, validation and test splits of D2 are disjoint by subject. When evaluating the model, we consider it's performance using entities that it has not been exposed to during training. \jt{@Marzieh: does the same hold for D1?}

\subsubsection*{Findings}

Table~\ref{table:spj} shows the EM of the intermediate results produced by the Neural SPJ (projection EM),  and the EM of the final answer (answer EM). 
In row 2 of Table~\ref{table:spj} we observe a substantial drop of more than 20\% in EM, both Projection, and Answer, when we add noise to the retrieved support set at test time. This suggests that adding noise from retrieval at training time (third row), makes the neural SPJ module more resilient to errors introduced by the SSG.% or IR module.

The Neural SPJ operator was also trained to predict the downstream aggregation function, which it did correctly with $99.97\%$ accuracy. In $0.03\%$ of cases, the wrong function name was selected. Answer EM reflects the accuracy of the end-to-end performance of the system. Furthermore, we inspected the error rate for each query type for the results reported in Table~\ref{table:spj} and noted that there was no large deviation between queries with aggregation and those without. This validates our choice to postpone aggregation functions to happen after the Neural SPJ.

\subsubsection*{Error Analysis}
For the Noisy IR model reported in Table~\ref{table:spj} were no false positive or false negative errors (the model outputs a result when it should output {\tt NULL} or vice versa). All errors were errors in the the content of the fact returned. Of the 253 errors, 166 were for symmetric relations (X borders Y) which introduced ambiguity in training. 80 errors were due to the model incorrectly applying implicit world knowledge when provided with insufficient facts to resolve geographical meronomy relations (e.g. incorrectly predicting Zeeland [sic] is in Asia). The remaining errors were differing punctuation (e.g. missing a period). 

\subsubsection*{Training data requirements}
\begin{figure}
\centering
    \includegraphics[width=0.8\linewidth]{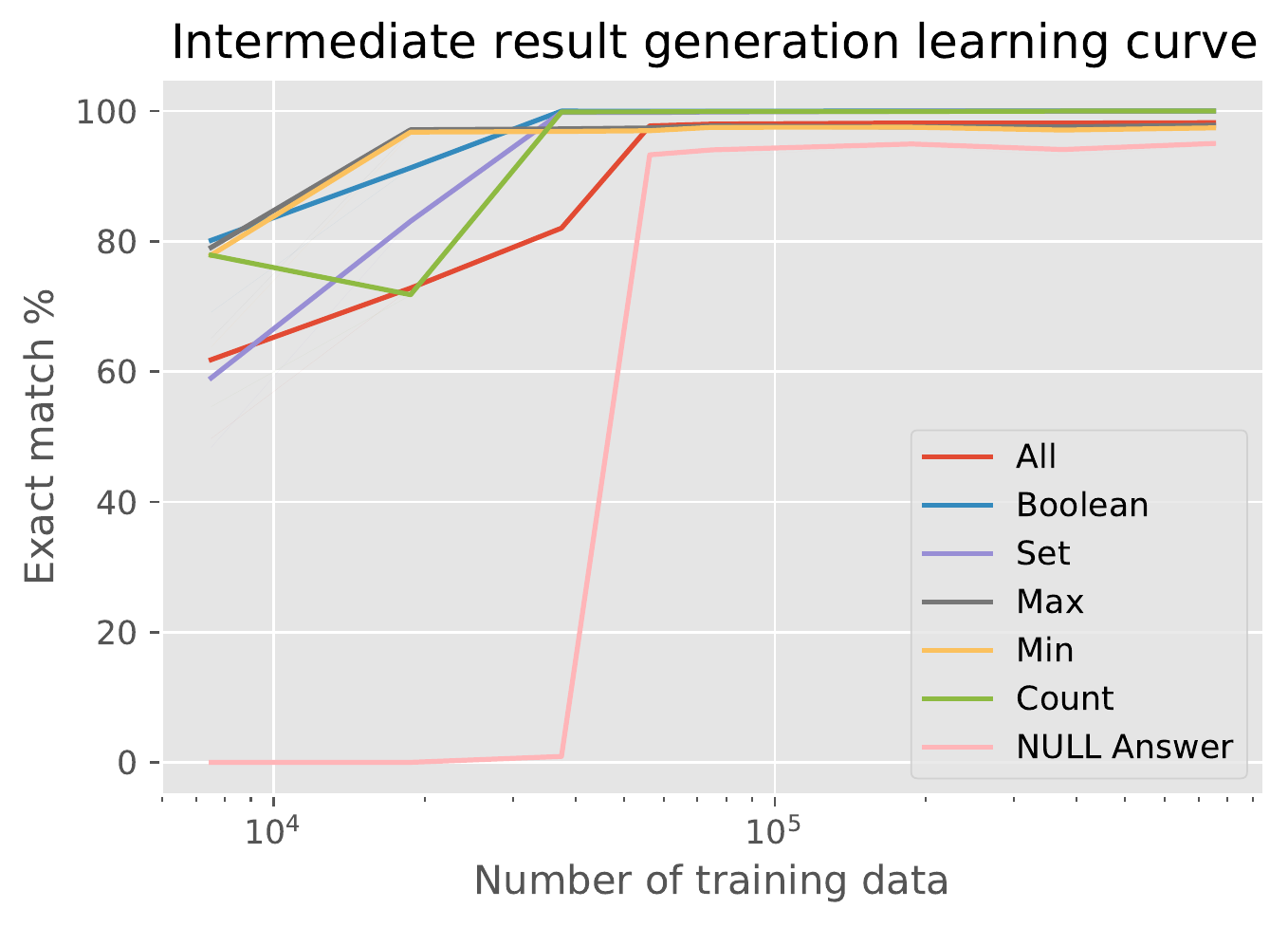}
    \caption{Training the Neural SPJ model with a varying number data indicates that 40k training instances are required to train a model and that learning when not to generate an answer (by predicting {\tt NULL}) can be dominated by other relations if there are insufficient examples for it.}
    \label{fig:results:sp-data}
\end{figure}

To evaluate training data requirements for the Neural SPJ, we plot the learning curve for generating intermediate results using dataset D2 in Figure~\ref{fig:results:sp-data}. For each point on this graph, 5 models are trained with different random initializations and we report the mean EM score. The model attains near perfect accuracy with approximately 80k training instances. It is interesting to note that the Neural SPJ struggled the most with learning what it {\em doesn't} know. With little training data, the model often outputted a result even when it should have outputted NULL. This is a well known weakness of sequence-to-sequence models \cite{Rajpurkar2018}.

% \subsubsection*{SPJ Breakdown and Error Analysis}
% We inspected the error rate for each query type for the results reported in Table~\ref{table:spj} and noted that there was no large deviation between queries with aggregation and those without. This validates our choice postpone aggregation functions to happen after the Neural SPJ.

% in performance from the reported EM score.
% We reported in Section~\ref{section:first-experiment}, that even with perfect IR, the transformer does not generate correct answers for some types of aggregation. We overcome this limitation by deferring aggregation as a separate computation over the output of the SPJ operator. Inspecting 

%Furthermore, the EM for queries requiring aggregations is consistently high while being able to scale to large databases, overcoming the encoder limitations within the transformer module. \fs{I can't parse the following sentence} For one of the 27 relations this model is trained on, EM is lower across multiple query types, a breakdown is provided in Section~\ref{section:error-analysis}.

\subsection{Support Set Generation}
\label{section:experiments:ssg}
In this section we evaluate how well the SSG component (Section~\ref{section:architecture:support-set-generation}) retrieves facts to be fed to the Neural SPJ operators. It is tricky to evaluate the SSG because errors in the SSG do not necessarily translate into errors in query answers. For example, the SSG may return a superset of a support set, but the Neural SPJ may still generate the correct answer. 

Table~\ref{supervised_issg} shows the performance of the SSG. In the table, an output of the SSG is considered an exact match (correspondingly, a soft match) if it is exactly the same as (correspondingly, a super set of) a support set in the reference data. The table shows the precision and recall of the distantly supervised SSG. The results for the supervised SSG are between 10-20\% higher and up to 30\% higher for queries with aggregation. We tuned the training to increase recall so more support sets are generated. 

As can be seen, while the results for lookups and joins are high, they drop significantly for aggregation and set answers. 
However, the point to note is that even in these cases, the SSG still learns what it needs. For example, consider training the SSG for the query {\em who is the oldest person?} When we train it by running a single Neural SPJ, it will rarely get the correct support set because that support set (any fact pertaining to people's age) is large and may not even be considered. However, the SSG still learns that it needs to find facts about people's ages. So when the SSG action classifier is applied to a single decision in Algorithm 2 (namely, {\em should I add a fact that is about a person's age to a support set?}), it will return True, thereby obtaining the desired result.

\begin{table}
\caption{Precision and recall of distantly supervised SSG w.r.t.\ the reference set. Note that errors in training do not necessarily translate to wrong query answers because the Neural SPJ operator is somewhat robust to extra information.}
\label{supervised_issg}
\label{ds_issg}
\label{table:ssg}
\begin{tabular}{@{}lcccc@{}}
\toprule
\multicolumn{1}{c}{\multirow{2}{*}{\textbf{\begin{tabular}[c]{@{}c@{}}Query \\ Type\end{tabular}}}} & \multicolumn{2}{c}{\textbf{\begin{tabular}[c]{@{}c@{}}Exact Match (\%)\end{tabular}}} & \multicolumn{2}{c}{\textbf{\begin{tabular}[c]{@{}c@{}}Soft Match (\%)\end{tabular}}} \\ \cmidrule(l){2-5} 
\multicolumn{1}{c}{}                                                                                & \multicolumn{1}{l}{\textbf{Precision}}          & \multicolumn{1}{l}{\textbf{Recall}}         & \textbf{Precision}                                & \textbf{Recall}                               \\ \midrule
Atomic (bool)                                                                                       & 81.45                                           & 98.92                                       & 95.02                                             & 99.90                                         \\
Atomic (extractive)                                                                                 & 80.44                                          & 88.47                                       & 94.00                                             & 99.61                                         \\ \midrule
Join (bool)                                                                                         & 62.28                                           & 83.13                                       & 62.28                                             & 83.13                                         \\
Join (extractive)                                                                                   & 72.22                                           & 85.95                                       & 72.22                                             & 85.95                                         \\ \midrule
Set                                                                                                 & 29.97                                           & 89.14                                       & 33.13                                             & 89.14                                         \\
Count                                                                                               & 56.42                                           & 92.80                                       & 59.68                                            & 92.80                                         \\
min/max                                                                                         & 68.63                                             & 100.00                                         & 68.63                                             & 100.00        
                                \\ \midrule
Total                                                                                                 & 68.28                                           & 93.57                                       & 77.25                                             & 95.68     
\\ \bottomrule
\end{tabular}
\end{table}

\subsection{Supervision and transfer learning}
\label{section:experiment-analysis}

\begin{figure}
\centering
    \includegraphics[width=0.8\linewidth]{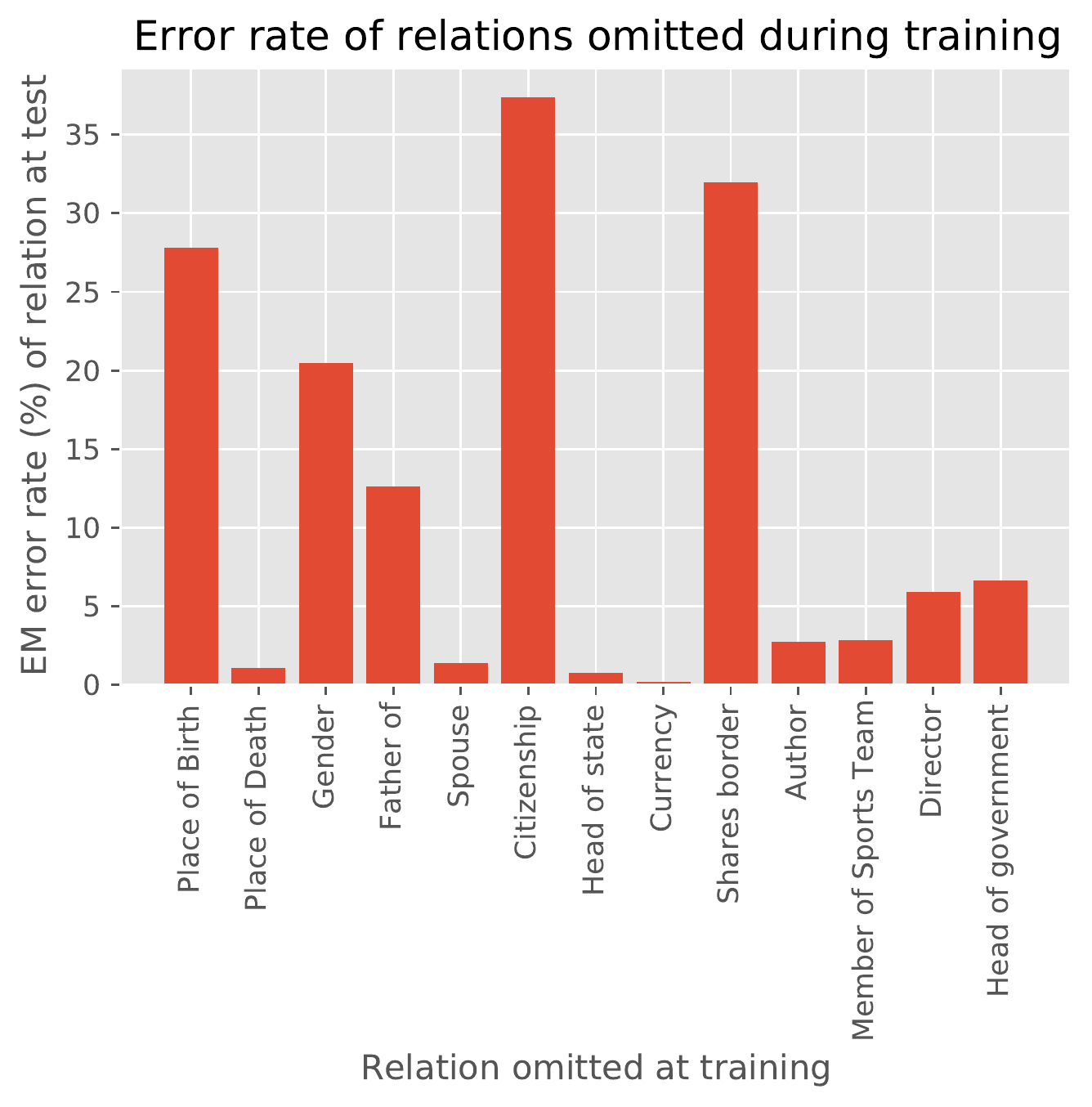}
    \caption{Even with relations omitted during training, Neural SPJ often generates correct intermediate results. }
    \label{fig:results:sp-trans}

\end{figure}

An important question for \ndb s is whether the system can answer queries about relationships it has not seen in the training data. We perform two experiments. In the first experiment we test the accuracy of answering queries on a single new relation that was not seen in training across a sample of relations. 
The result is shown in Figure~\ref{fig:results:sp-trans} where the bars indicate the error rate at test time for a model with that relation omitted during training.
%The result is shown in Figure~\ref{fig:results:sp-trans} where the leading diagonal shows an increase in error rate for the relation omitted during training.

In the second experiment, we double the number of relations in the test set compared to the training set. This corresponds to a scenario in which there is a massive sudden shift in the queries that highlights the importance of the underlying language model.  When we trained on 13 out of 27 relations and evaluated on all 27, EM falls from 99\% to 86\%.  However, if we started with a randomly initialized language model and trained on the same 13 relations, then our accuracy drops to only 55\% EM, showing that the pre-trained language model is critical to the performance of the \ndb.

\section{Related Work}

\subsubsection*{NLP and data management}
Bridging the gap between unstructured natural language data and database-style querying has been a long-standing theme in database research~\cite{DBLP:conf/cidr/HalevyED03}. 
The work on information extraction has developed techniques for translating segments of natural language text into triples that can be further processed by a database system. 
Wikidata~\cite{vrandevcic2014wikidata} itself is a social experiment where additions to the knowledge graph are encouraged to use already existing relation names if possible, thereby alleviating the need for information extraction. 
There has been significant work on translating queries posed in natural language into SQL queries on a database whose schema is known~\cite{Androutsopoulos1995, Li2014,DBLP:journals/corr/abs-2007-15280}, with extensions to semi-structured data and knowledge bases~\cite{Pasupat2015,Berant2013}. More recently, systems such as {BREAK}~\cite{Wolfson2020} and {ShARC}~\cite{saeidi2018interpretation} have trained models to translate a natural language query into a sequence of relational operators (or variants thereof).

\ndb s do not try to map data or queries into a pre-defined schema. At the core, we use neural techniques to process the facts in the database with the query given as context in natural language. However, \ndb s do some rudimentary analysis of the query when they decide whether it requires an aggregation operator, and one can imagine that \ndb s will need more sophisticated understanding of the structure of a query as they tackle more complex queries. Similarly, the processing performed by the Neural SPJ operator is reminiscent of information extraction in the sense that it produces a structured representation of facts that can be used by subsequent operators. However, a key difference is that the extraction performed by the Neural SPJ is query dependent and is independent of any schema. 

With similar goals in mind, the Information Retrieval community has developed search engines to answer SQL queries~\cite{fuhr1996models, AmerYahia05report}. The work most close to ours~\cite{weikum2007db}, explores the problem of answering queries from a collection of non-schematic XML documents that exhibit  heterogeneous structures, and hence are cumbersome in languages such as XPath or XQuery. Another similar research line is that of Whang \emph{et al.}~\cite{whang2013odys, whang2015db}. Similarly to what we propose they also support natural language queries but they still exploit semi-structured data. % to answer queries such as: ``Finds papers about “cloud computing” published after “2005” in digital libraries''. 
Whereas in our case, %to solve this query 
the system needs to ``understand'' what are the relations and attributes that need to be used and the relative operators %.and at the same time IR techniques are used to retrieve the initial set of potentially relevant information.

\subsubsection*{Question answering from text}
The NLP community has made great strides recently on the problem of answering queries from text, which includes tasks such as open-book question answering~\cite{Richardson2013, Rajpurkar2016} and fact verification~\cite{Thorne2018a}. To efficiently scale machine comprehension to very large databases, the NLP community adopt either a pipelined~\cite{Thorne2018a, Lewis2020} or jointly trained~\cite{Guu2020} architecture of information retrieval with neural reasoning.
Like \ndb s, many of these works answer questions using an explicit memory of knowledge (free-form text) in addition to the pre-trained language model~\cite{Lewis2020,petroni2020kilt,Guu2020}. 
However, these works typically require extracting a span from a single document or predicting a token or label as an answer, whereas \ndb s require combining multiple facts, performing selections and aggregation. 
While in-roads have been made to perform discrete reasoning over passages \cite{Dua2019}, with explicit computation \cite{Andor2019}, these use only a single passage rather than requiring aggregation over large numbers of facts.

Multi-hop question answering is a recent setting where answering a query requires finding supporting evidence in multiple documents (see \cite{welbl2018constructing,Talmor2018,Wolfson2020} for data sets facilitating this research). In solving multi-hop questions, the works either decompose the question into simpler sub questions~\cite{min2019multi,Wolfson2020}, or condition each hop on the previously retrieved documents~\cite{asai2019learning}. Some of these ideas inspired the design of the SSG in Section~\ref{section:architecture:support-set-generation}.  Transformers have been shown to perform soft-reasoning when provided with simple propositional rules~\cite{Clark2020}. In that work, transformers were able to join a small number of facts and rules of the form $A \rightarrow B$.

While other works modeling the web as a knowledge bases have focused on combining multiple snippets of text together \cite{Talmor2018}, their assumption is that the query is decomposed into a SPARQL program that is executed on pre-extracted information.
Our innovation is that no latent program or structure is needed and that information extraction is dynamic and dependent on the query.

%%%%%%%%%%%%%%%%%%%%%%%%%%%%%%%%% IR + DB perspective %%%%%%%%%%%%%%%%%%%%%%

\subsubsection*{Extending neural architectures to reasoning tasks}

In the same spirit to Neural Turing Machines\cite{DBLP:journals/corr/GravesWD14} and Memory Networks\cite{sukhbaatar2015end} architectures, an alternative way of building \ndb\ is to encode all the facts in the database to a neural memory and build machinery to read, write, and reason on top of this neural memory. However, such an approach would not have control and transparency: It is challenging to remove facts from the database or check whether a particular fact exists.  Also, it would not be possible to explain query results. Furthermore, these architectures perform well on bAbI \cite{weston2015towards} tasks where the number of facts is limited, and mainly lookup or simple reasoning is needed. But, In our experiments in \ndb\, they couldn't perform well; we hypothesize that encoding the query and facts together by a stack of self-attention in the encoder is necessary to answer database queries. 
There also have been considerable efforts in mixing traditional symbolic reasoning or data management algorithms with neural network architectures. For example, Rockt\"{a}schel \emph{et al. }~\cite{Rocktaschel2017} have developed a differentiable version of the backward chaining algorithm that drives prolog. Most closely to our work, Minervini \emph{et al.}~\cite{Minervini2020} has showed how differentiable prolog interpreters can be used to support reasoning with facts in natural language. Instead of ``neuralizing'' existing symbolic reasoners, in our work we start off with a scalable neural architecture, and support it with symbolic computation only where necessary. This enables us to directly leverage the rapid progress made in retrieval augmented QA models and ensures scalability.

\section{Conclusions and future work}
\label{section:conclusions}

We described \ndb, a new kind of database system that uses neural reasoning, and is therefore able to answer queries from data expressed as natural language sentences that do not conform to a pre-defined schema. The design of the \ndb\ architecture was based on a careful examination of the strengths and weaknesses of current NLP transformer models. Our experimental results suggest that it is possible to attain very high accuracy for a class of queries that involve select, project, join possibly followed by an aggregation.

To fully realize the promise of \ndb s, more research is needed on scaling up \ndb s to larger databases, supporting more complex queries and increasing the accuracy of the answers. In particular, an interesting area of research noted in Section~\ref{section:architecture:support-set-generation} is developing novel indexing techniques that enable efficient support set generation.  Another exciting area to investigate is to consider other media in the database. For example, a database can also contain a set of images and some queries can involve combining information from language and from images. Such an extension would benefit from recent progress on visual query answering systems~\cite{DBLP:conf/iccv/AntolALMBZP15,DBLP:conf/cvpr/AndreasRDK16}. 

A possible downside of using neural techniques in a database system is the potential for bias that might be encoded in the underlying language model. As we discussed, the fact that London in the UK is encoded in the language model and is useful.  However, suppose our database included facts saying that John and Jane work at a hospital, but when we asked what their profession is, the system would answer doctor for John and nurse for Jane. Currently, there is no good way of distinguishing biased from unbiased knowledge in a language model. A possible approach to this important issue is to design a separate module that attacks the database with queries in order to discover hidden biases. Then, we could devise safeguards within the database that ensure that we don't use such biased knowledge in answering queries. Developing these components is an area for future research. 

Finally, another interesting challenge concerns developing semantic knowledge that helps in identifying which updates should replace previous facts and which should not. For example, if the fact, {\sf Mariah is unemployed}, was in the database and later the fact, {\sf Mariah works for Apple}, was added, then the first fact should be removed (or at least, apply only to queries about the past). However, the same does not hold for the facts {\sf Kasper likes tea} followed by the fact {\sf Kasper likes coffee}.

\bibliographystyle{plain}
\bibliography{references}

\begin{thebibliography}{10}

\bibitem{AmerYahia05report}
Sihem Amer-Yahia, Pat Case, Thomas R\"{o}lleke, Jayavel Shanmugasundaram, and
  Gerhard Weikum.
\newblock Report on the db/ir panel at sigmod 2005.
\newblock {\em SIGMOD Rec.}, 34(4):71–74, December 2005.

\bibitem{Andor2019}
Daniel Andor, Luheng He, Kenton Lee, and Emily Pitler.
\newblock {Giving Bert a calculator: Finding operations and arguments with
  reading comprehension}.
\newblock {\em EMNLP-IJCNLP 2019 - 2019 Conference on Empirical Methods in
  Natural Language Processing and 9th International Joint Conference on Natural
  Language Processing, Proceedings of the Conference}, 2:5947--5952, 2019.

\bibitem{DBLP:conf/cvpr/AndreasRDK16}
Jacob Andreas, Marcus Rohrbach, Trevor Darrell, and Dan Klein.
\newblock Neural module networks.
\newblock In {\em 2016 {IEEE} Conference on Computer Vision and Pattern
  Recognition, {CVPR} 2016, Las Vegas, NV, USA, June 27-30, 2016}, pages
  39--48. {IEEE} Computer Society, 2016.

\bibitem{Androutsopoulos1995}
I~Androutsopoulos, G~D Ritchie, and P~Thanisch.
\newblock {Natural Language Interfaces to Databases - an Introduction}.
\newblock {\em Natural Language Engineering}, 1(1):29--81, 1995.

\bibitem{DBLP:conf/iccv/AntolALMBZP15}
Stanislaw Antol, Aishwarya Agrawal, Jiasen Lu, Margaret Mitchell, Dhruv Batra,
  C.~Lawrence Zitnick, and Devi Parikh.
\newblock {VQA:} visual question answering.
\newblock In {\em 2015 {IEEE} International Conference on Computer Vision,
  {ICCV} 2015, Santiago, Chile, December 7-13, 2015}, pages 2425--2433. {IEEE}
  Computer Society, 2015.

\bibitem{asai2019learning}
Akari Asai, Kazuma Hashimoto, Hannaneh Hajishirzi, Richard Socher, and Caiming
  Xiong.
\newblock Learning to retrieve reasoning paths over wikipedia graph for
  question answering.
\newblock {\em arXiv preprint arXiv:1911.10470}, 2019.

\bibitem{Berant2013}
Jonathan Berant, Andrew Chou, Roy Frostig, and Percy Liang.
\newblock {Semantic parsing on freebase from question-answer pairs}.
\newblock {\em EMNLP 2013 - 2013 Conference on Empirical Methods in Natural
  Language Processing, Proceedings of the Conference}, (October):1533--1544,
  2013.

\bibitem{Brown2020}
Tom~B Brown, Benjamin Mann, Nick Ryder, Melanie Subbiah, Jared Kaplan, Prafulla
  Dhariwal, Arvind Neelakantan, Pranav Shyam, Girish Sastry, Amanda Askell,
  Sandhini Agarwal, Ariel Herbert-Voss, Gretchen Krueger, Tom Henighan, Rewon
  Child, Aditya Ramesh, Daniel~M Ziegler, Jeffrey Wu, Clemens Winter,
  Christopher Hesse, Mark Chen, Eric Sigler, Mateusz Litwin, Scott Gray,
  Benjamin Chess, Jack Clark, Christopher Berner, Sam McCandlish, Alec Radford,
  Ilya Sutskever, and Dario Amodei.
\newblock {Language Models are Few-Shot Learners}.
\newblock 2020.

\bibitem{Clark2020}
Peter Clark, Oyvind Tafjord, and Kyle Richardson.
\newblock {Transformers as Soft Reasoners over Language}.
\newblock {\em IJCAI}, pages 3882--3890, 2020.

\bibitem{Devlin2019}
Jacob Devlin, Ming-Wei Chang, Kenton Lee, and Kristina Toutanova.
\newblock {BERT: Pre-training of Deep Bidirectional Transformers for Language
  Understanding}.
\newblock In {\em Proceedings of the 2019 Conference of the North American
  Chapter of the Association for Computational Linguistics: Human Language
  Technologies, Volume 1 (Long and Short Papers)}, Minneapolis, Minnesota,
  2019.

\bibitem{Dua2019}
Dheeru Dua, Yizhong Wang, Pradeep Dasigi, Gabriel Stanovsky, Sameer Singh, and
  Matt Gardner.
\newblock {{DROP}: A Reading Comprehension Benchmark Requiring Discrete
  Reasoning Over Paragraphs}.
\newblock In {\em Proceedings of the 2019 Conference of the North {A}merican
  Chapter of the Association for Computational Linguistics: Human Language
  Technologies, Volume 1 (Long and Short Papers)}, pages 2368--2378,
  Minneapolis, Minnesota, jun 2019. Association for Computational Linguistics.

\bibitem{fuhr1996models}
Norbert Fuhr.
\newblock Models for integrated information retrieval and database systems.
\newblock {\em IEEE Data Eng. Bull.}, 19(1):3--13, 1996.

\bibitem{DBLP:journals/corr/GravesWD14}
Alex Graves, Greg Wayne, and Ivo Danihelka.
\newblock Neural turing machines.
\newblock {\em CoRR}, abs/1410.5401, 2014.

\bibitem{Guu2020}
Kelvin Guu, Kenton Lee, Zora Tung, Panupong Pasupat, and Ming-wei Chang.
\newblock {REALM : Retrieval-Augmented Language Model Pre-Training}.
\newblock 2020.

\bibitem{DBLP:conf/cidr/HalevyED03}
Alon~Y. Halevy, Oren Etzioni, AnHai Doan, Zachary~G. Ives, Jayant Madhavan,
  Luke~K. McDowell, and Igor Tatarinov.
\newblock Crossing the structure chasm.
\newblock In {\em {CIDR} 2003, First Biennial Conference on Innovative Data
  Systems Research, Asilomar, CA, USA, January 5-8, 2003, Online Proceedings}.
  www.cidrdb.org, 2003.

\bibitem{Hao2019}
Yaru Hao, Li~Dong, Furu Wei, and Ke~Xu.
\newblock {Visualizing and understanding the effectiveness of BERT}.
\newblock {\em EMNLP-IJCNLP 2019 - 2019 Conference on Empirical Methods in
  Natural Language Processing and 9th International Joint Conference on Natural
  Language Processing, Proceedings of the Conference}, pages 4143--4152, 2019.

\bibitem{Hupkes2020}
Dieuwke Hupkes, Verna Dankers, Mathijs Mul, and Elia Bruni.
\newblock {Compositionality Decomposed: How do Neural Networks Generalise?}
\newblock {\em Journal of Artificial Intelligence Research}, 67:757--795, 2020.

\bibitem{Izacard2020}
Gautier Izacard and Edouard Grave.
\newblock {Leveraging Passage Retrieval with Generative Models for Open Domain
  Question Answering}.
\newblock 2020.

\bibitem{Johnson2019}
Jeff Johnson, Matthijs Douze, and Herve Jegou.
\newblock {Billion-scale similarity search with GPUs}.
\newblock {\em IEEE Transactions on Big Data}, pages 1--1, 2019.

\bibitem{Karpukhin2020}
Vladimir Karpukhin, Barlas Oğuz, Sewon Min, Patrick Lewis, Ledell Wu, Sergey
  Edunov, Danqi Chen, and Wen-tau Yih.
\newblock {Dense Passage Retrieval for Open-Domain Question Answering}.
\newblock 2020.

\bibitem{DBLP:conf/sigmod/KraskaBCDP18}
Tim Kraska, Alex Beutel, Ed~H. Chi, Jeffrey Dean, and Neoklis Polyzotis.
\newblock The case for learned index structures.
\newblock In Gautam Das, Christopher~M. Jermaine, and Philip~A. Bernstein,
  editors, {\em Proceedings of the 2018 International Conference on Management
  of Data, {SIGMOD} Conference 2018, Houston, TX, USA, June 10-15, 2018}, pages
  489--504. {ACM}, 2018.

\bibitem{Lewis2020}
Patrick Lewis, Ethan Perez, Aleksandara Piktus, Fabio Petroni, Vladimir
  Karpukhin, Naman Goyal, Heinrich K{\"{u}}ttler, Mike Lewis, Wen-tau Yih, Tim
  Rockt{\"{a}}schel, Sebastian Riedel, and Douwe Kiela.
\newblock {Retrieval-Augmented Generation for Knowledge-Intensive NLP Tasks}.
\newblock 2020.

\bibitem{Li2014}
Fei Li and H~V Jagadish.
\newblock {Constructing an Interactive Natural Language Interface for
  Relational Databases}.
\newblock {\em Proceedings of the VLDB Endowment2}, 8(1):73--84, 2014.

\bibitem{Li2016}
Jiwei Li, Will Monroe, and Dan Jurafsky.
\newblock {Understanding Neural Networks through Representation Erasure}.
\newblock 2016.

\bibitem{DBLP:journals/corr/abs-2004-00584}
Yuliang Li, Jinfeng Li, Yoshihiko Suhara, AnHai Doan, and Wang{-}Chiew Tan.
\newblock Deep entity matching with pre-trained language models.
\newblock {\em CoRR}, abs/2004.00584, 2020.

\bibitem{Liu2019RobustlyOptimized}
Yinhan Liu, Myle Ott, Naman Goyal, Jingfei Du, Mandar Joshi, Danqi Chen, Omer
  Levy, Mike Lewis, Luke Zettlemoyer, and Veselin Stoyanov.
\newblock {RoBERTa: A Robustly Optimized BERT Pretraining Approach}.
\newblock 2019.

\bibitem{min2019multi}
Sewon Min, Victor Zhong, Luke Zettlemoyer, and Hannaneh Hajishirzi.
\newblock Multi-hop reading comprehension through question decomposition and
  rescoring.
\newblock {\em arXiv preprint arXiv:1906.02916}, 2019.

\bibitem{Minervini2020}
Pasquale Minervini, Matko Bosnjak, Tim Rockt{\"{a}}schel, Sebastian Riedel, and
  Edward Grefenstette.
\newblock Differentiable reasoning on large knowledge bases and natural
  language.
\newblock In {\em The Thirty-Fourth {AAAI} Conference on Artificial
  Intelligence, {AAAI} 2020, The Thirty-Second Innovative Applications of
  Artificial Intelligence Conference, {IAAI} 2020, The Tenth {AAAI} Symposium
  on Educational Advances in Artificial Intelligence, {EAAI} 2020, New York,
  NY, USA, February 7-12, 2020}, pages 5182--5190. {AAAI} Press, 2020.

\bibitem{DBLP:conf/sigmod/MudgalLRDPKDAR18}
Sidharth Mudgal, Han Li, Theodoros Rekatsinas, AnHai Doan, Youngchoon Park,
  Ganesh Krishnan, Rohit Deep, Esteban Arcaute, and Vijay Raghavendra.
\newblock Deep learning for entity matching: {A} design space exploration.
\newblock In Gautam Das, Christopher~M. Jermaine, and Philip~A. Bernstein,
  editors, {\em Proceedings of the 2018 International Conference on Management
  of Data, {SIGMOD} Conference 2018, Houston, TX, USA, June 10-15, 2018}, pages
  19--34. {ACM}, 2018.

\bibitem{Pasupat2015}
Panupong Pasupat and Percy Liang.
\newblock {Compositional Semantic Parsing on Semi-Structured Tables}.
\newblock {\em Proceedings of the 53rd Annual Meeting of the Association for
  Computational Linguistics and the 7th International Joint Conference on
  Natural Language Processing (Volume 1: Long Papers)}, pages 1470--1480, 2015.

\bibitem{Peters2018}
Matthew Peters, Mark Neumann, Luke Zettlemoyer, and Wen-tau Yih.
\newblock {Dissecting Contextual Word Embeddings: Architecture and
  Representation}.
\newblock In {\em Proceedings of the 2018 Conference on Empirical Methods in
  Natural Language Processing}, pages 1499--1509, Brussels, Belgium, 2018.
  Association for Computational Linguistics.

\bibitem{petroni2020kilt}
Fabio Petroni, Aleksandra Piktus, Angela Fan, Patrick Lewis, Majid Yazdani,
  Nicola De~Cao, James Thorne, Yacine Jernite, Vassilis Plachouras, Tim
  Rockt{\"a}schel, et~al.
\newblock Kilt: a benchmark for knowledge intensive language tasks.
\newblock {\em arXiv preprint arXiv:2009.02252}, 2020.

\bibitem{Petroni2019}
Fabio Petroni, Tim Rockt{\"{a}}schel, Patrick Lewis, Anton Bakhtin, Yuxiang Wu,
  Alexander~H. Miller, and Sebastian Riedel.
\newblock {Language Models as Knowledge Bases?}
\newblock In {\em Proceedings of EMNLP-IJCNLP}, Hong Kong, China, 2019.
  Association for Computational Linguistics.

\bibitem{Radford2018}
Alec Radford, Karthik Narasimhan, Tim Salimans, and Ilya Sutskever.
\newblock {Improving Language Understanding by Generative Pre-Training}.
\newblock {\em arXiv}, pages 1--12, 2018.

\bibitem{Raffel2020}
Colin Raffel, Noam Shazeer, Adam Roberts, Katherine Lee, Sharan Narang, Michael
  Matena, Yanqi Zhou, Wei Li, and Peter~J. Liu.
\newblock {Exploring the Limits of Transfer Learning with a Unified
  Text-to-Text Transformer}.
\newblock {\em Journal of Machine Learning Research}, 21:1--67, 2020.

\bibitem{Rajpurkar2018}
Pranav Rajpurkar, Robin Jia, and Percy Liang.
\newblock {Know what you don't know: Unanswerable questions for SQuAD}.
\newblock {\em ACL 2018 - 56th Annual Meeting of the Association for
  Computational Linguistics, Proceedings of the Conference (Long Papers)},
  2:784--789, 2018.

\bibitem{Rajpurkar2016}
Pranav Rajpurkar, Jian Zhang, Konstantin Lopyrev, and Percy Liang.
\newblock {SQuAD: 100,000+ Questions for Machine Comprehension of Text}.
\newblock pages 2383--2392, 2016.

\bibitem{Ribeiro2016}
Marco~Tulio Ribeiro, Sameer Singh, and Carlos Guestrin.
\newblock {"Why Should I Trust You?": Explaining the Predictions of Any
  Classifier}.
\newblock 39(2011):117831, 2016.

\bibitem{Richardson2013}
Matthew Richardson, Christopher J~C Burges, and Erin Renshaw.
\newblock {{\{}MCT{\}}est: A Challenge Dataset for the Open-Domain Machine
  Comprehension of Text}.
\newblock In {\em Proceedings of the 2013 Conference on Empirical Methods in
  Natural Language Processing}, pages 193--203, Seattle, Washington, USA, oct
  2013. Association for Computational Linguistics.

\bibitem{Rocktaschel2017}
Tim Rockt{\"{a}}schel and Sebastian Riedel.
\newblock {End-to-end Differentiable Proving}.
\newblock In I~Guyon, U~V Luxburg, S~Bengio, H~Wallach, R~Fergus,
  S~Vishwanathan, and R~Garnett, editors, {\em Advances in Neural Information
  Processing Systems 30}, pages 3788--3800. Curran Associates, Inc., 2017.

\bibitem{saeidi2018interpretation}
Marzieh Saeidi, Max Bartolo, Patrick Lewis, Sameer Singh, Tim Rockt{\"a}schel,
  Mike Sheldon, Guillaume Bouchard, and Sebastian Riedel.
\newblock Interpretation of natural language rules in conversational machine
  reading.
\newblock {\em arXiv preprint arXiv:1809.01494}, 2018.

\bibitem{Stickland2019}
Asa~Cooper Stickland and Iain Murray.
\newblock {BERT and PALs: Projected attention layers for efficient adaptation
  in multi-task learning}.
\newblock {\em 36th International Conference on Machine Learning, ICML 2019},
  2019-June:10477--10488, 2019.

\bibitem{sukhbaatar2015end}
Sainbayar Sukhbaatar, Jason Weston, Rob Fergus, et~al.
\newblock End-to-end memory networks.
\newblock In {\em Advances in neural information processing systems}, pages
  2440--2448, 2015.

\bibitem{Talmor2018}
Alon Talmor and Jonathan Berant.
\newblock {The web as a knowledge-base for answering complex questions}.
\newblock {\em NAACL HLT 2018 - 2018 Conference of the North American Chapter
  of the Association for Computational Linguistics: Human Language Technologies
  - Proceedings of the Conference}, 1:641--651, 2018.

\bibitem{Tenney2019}
Ian Tenney, Patrick Xia, Berlin Chen, Alex Wang, Adam Poliak, R.~Thomas McCoy,
  Najoung Kim, Benjamin~Van Durme, Samuel~R Bowman, Dipanjan Das, and Ellie
  Pavlick.
\newblock {What do you learn from context? Probing for sentence structure in
  contextualized word representations}.
\newblock {\em ICLR}, pages 1--17, 2019.

\bibitem{Thorne2018a}
James Thorne, Andreas Vlachos, Christos Christodoulopoulos, and Arpit Mittal.
\newblock {FEVER: a large-scale dataset for Fact Extraction and VERification}.
\newblock In {\em Proceedings of the 2018 Conference of the North American
  Chapter of the Association for Computational Linguistics: Human Language
  Technologies, Volume 1 (Long Papers)}, pages 809--819, New Orleans,
  Louisiana, 2018. Association for Computational Linguistics.

\bibitem{Vaswani2017}
Ashish Vaswani, Noam Shazeer, Niki Parmar, Jakob Uszkoreit, Lilon Jones, Aidan
  Gomez, {\L}ukasz Kaiser, and Illia Polosukhin.
\newblock {Attention is all you need}.
\newblock In {\em 31st Conference on Neural Information Processing Systems
  (NIPS 2017)}, Long Beach, CA, USA, 2017.

\bibitem{vrandevcic2014wikidata}
Denny Vrande{\v{c}}i{\'c} and Markus Kr{\"o}tzsch.
\newblock Wikidata: a free collaborative knowledgebase.
\newblock {\em Communications of the ACM}, 57(10):78--85, 2014.

\bibitem{Wang2020}
Sinong Wang, Belinda~Z. Li, Madian Khabsa, Han Fang, and Hao Ma.
\newblock {Linformer: Self-Attention with Linear Complexity}.
\newblock 2048(2019), 2020.

\bibitem{Watkins1992}
Christopher J C~H Watkins and Peter Dayan.
\newblock {Q-learning}.
\newblock {\em Machine Learning}, 8(3):279--292, 1992.

\bibitem{weikum2007db}
Gerhard Weikum.
\newblock Db\&ir: both sides now.
\newblock In {\em Proceedings of the 2007 ACM SIGMOD international conference
  on Management of data}, pages 25--30, 2007.

\bibitem{welbl2018constructing}
Johannes Welbl, Pontus Stenetorp, and Sebastian Riedel.
\newblock Constructing datasets for multi-hop reading comprehension across
  documents.
\newblock {\em Transactions of the Association for Computational Linguistics},
  6:287--302, 2018.

\bibitem{weston2015towards}
Jason Weston, Antoine Bordes, Sumit Chopra, Alexander~M Rush, Bart van
  Merri{\"e}nboer, Armand Joulin, and Tomas Mikolov.
\newblock Towards ai-complete question answering: A set of prerequisite toy
  tasks.
\newblock {\em arXiv preprint arXiv:1502.05698}, 2015.

\bibitem{whang2015db}
Kyu-Young Whang, Jae-Gil Lee, Min-Jae Lee, Wook-Shin Han, Min-Soo Kim, and
  Jun-Sung Kim.
\newblock Db-ir integration using tight-coupling in the odysseus dbms.
\newblock {\em World Wide Web}, 18(3):491--520, 2015.

\bibitem{whang2013odys}
Kyu-Young Whang, Tae-Seob Yun, Yeon-Mi Yeo, Il-Yeol Song, Hyuk-Yoon Kwon, and
  In-Joong Kim.
\newblock Odys: an approach to building a massively-parallel search engine
  using a db-ir tightly-integrated parallel dbms for higher-level
  functionality.
\newblock In {\em Proceedings of the 2013 ACM SIGMOD International Conference
  on Management of Data}, pages 313--324, 2013.

\bibitem{wolf2020huggingfaces}
Thomas Wolf, Lysandre Debut, Victor Sanh, Julien Chaumond, Clement Delangue,
  Anthony Moi, Pierric Cistac, Tim Rault, Rémi Louf, Morgan Funtowicz, Joe
  Davison, Sam Shleifer, Patrick von Platen, Clara Ma, Yacine Jernite, Julien
  Plu, Canwen Xu, Teven~Le Scao, Sylvain Gugger, Mariama Drame, Quentin Lhoest,
  and Alexander~M. Rush.
\newblock Huggingface's transformers: State-of-the-art natural language
  processing, 2020.

\bibitem{Wolfson2020}
Tomer Wolfson, Mor Geva, Ankit Gupta, Matt Gardner, Yoav Goldberg, Daniel
  Deutch, and Jonathan Berant.
\newblock {Break It Down: A Question Understanding Benchmark}.
\newblock {\em Transactions of the Association for Computational Linguistics},
  8:183--198, 2020.

\bibitem{DBLP:journals/corr/abs-2007-15280}
Jichuan Zeng, Xi~Victoria Lin, Caiming Xiong, Richard Socher, Michael~R. Lyu,
  Irwin King, and Steven C.~H. Hoi.
\newblock Photon: {A} robust cross-domain text-to-sql system.
\newblock {\em CoRR}, abs/2007.15280, 2020.

\end{thebibliography}
\end{document}